\documentclass[runningheads]{llncs}

\usepackage{eccv}

\usepackage{eccvabbrv}

\usepackage{graphicx}
\usepackage{booktabs}
\usepackage{boldline}
\usepackage{manfnt}
\usepackage{wasysym}
\usepackage{amssymb}
\usepackage{makecell}
\usepackage{multicol}
\usepackage{multirow}
\usepackage[table]{xcolor}

\usepackage[accsupp]{axessibility}  

\usepackage{hyperref}

\usepackage{orcidlink}

\begin{document}

\title{AVTok: 1D Unified Tokenization for Holistic Audio-Video Generation} 


\author{Kien T. Pham\inst{1}\orcidlink{0009-0003-4529-8584} \and
I Chieh Chen\inst{1}\orcidlink{0009-0006-8999-7979} \and
Qifeng Chen\inst{1}\orcidlink{0000-0003-2199-3948} \and
Long Chen\inst{1}\thanks{Corresponding author.}\orcidlink{0000-0001-6148-9709}}

\authorrunning{K. Pham et al.}

\institute{The Hong Kong University of Science and Technology\\
\email{\{tkpham,icchen\}@connect.ust.hk, \{cqf,longchen\}@ust.hk}\\
Project Page: \url{https://hkust-longgroup.github.io/AVTok/}
}

\newcommand{\lc}[1]{{\color{blue}{$^\textbf{\emph{Long:}}$[#1]}}}

\maketitle

\begin{figure*}
  \centering
  \includegraphics[width=1.0\linewidth]{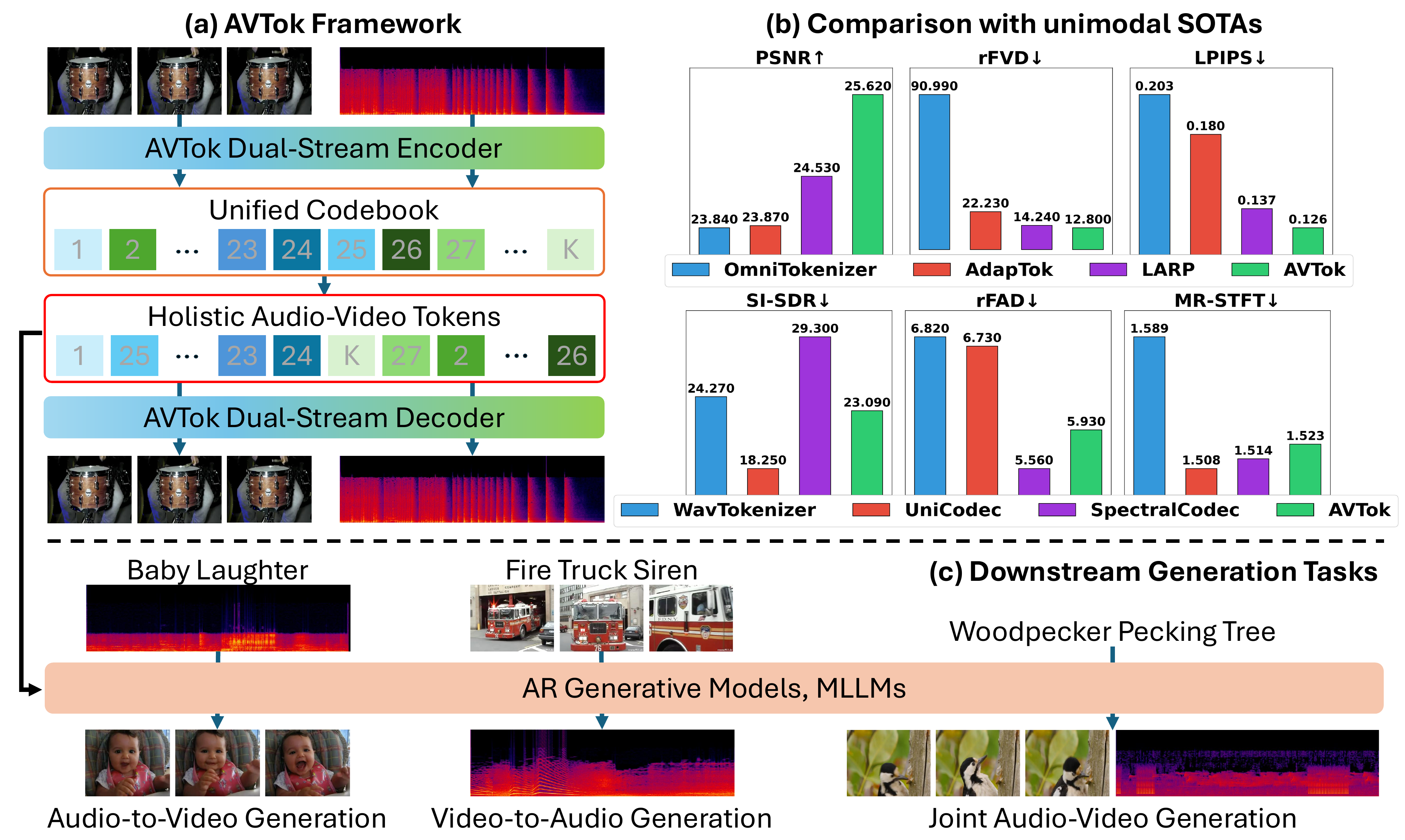}
  \caption{\textbf{Highlights.} \textbf{(a)} We propose AVTok, a novel unified tokenizer with dual-stream transformer-based architecture, capable of jointly encoding an audio-video pair into a single compact 1D latent representation. \textbf{(b)} AVTok achieves competitive performance compared to state-of-the-art unimodal 1D video tokenizers (\emph{top}) and audio codecs (\emph{bottom}). \textbf{(c)} From left to right, AVTok can be integrated into AR generative models to achieve audio-to-video, video-to-audio, and joint audio-video generation.}
  \label{fig:teaser}
\end{figure*}

\begin{abstract}
Audio-video generation has recently gained unprecedented research attention, aiming to synthesize high-quality sounding video content with fine-grained synchronization and semantic alignment between the auditory and visual components. The preceding methods predominantly adopt a dual-branch design with separate tokenization and generation modules per modality, neglecting the representation gap while necessitating intensive computational resources for proper training. Inspired by recent advancements in one-dimensional visual tokenization, we present \textbf{AVTok}, a novel unified tokenizer designated for holistic audio-video generation. AVTok features a dual-stream transformer-based architecture with shared encoder-decoder and modal-specific learnable queries to efficiently and effectively encode an audio-video pair into a compact one-dimensional latent representation with a unified codebook. To cope with the heterogeneous information imbalance that hinders AVTok from exploiting aligned audio-visual information, we devise a hierarchical training strategy to progressively realize reconstruction capabilities for each modality. Extensive experiments demonstrate that AVTok excels both in audio-video reconstruction and when integrated into downstream pipelines for audio-to-video, video-to-audio, and class-conditional joint audio-video generation. AVTok paves the way for the challenge of joint audio-video tokenization and provides a potential direction to build unified large multimodal models for audio-video generation.
  
\end{abstract}

\section{Introduction}
\label{sec:intro}
Audio-Visual (AV) content creation has undergone a remarkable transformation in recent years, catalyzing the emergence of innovative creative tasks that were once considered unattainable. This evolution has been largely driven by the development of powerful generative models, which are capable of Video-to-Audio (V2A)~\cite{mmaudio,v-aura,vintage,specvqgan,foleycrafter}, Audio-to-Video (A2V)~\cite{tempotoken,seeing-and-hearing,spa2v,weng2026audiosync,song2026syncphony}, and particularly Joint Audio-Video generation (JAVG)~\cite{ovi,javisdit,uniavgen,rflav,wan2p2,seed1p5pro,zheng2026aligning}. However, their impressive performance comes with a great price. These AV models typically adopt a heavy-weighted dual-branch architecture in which each processes one specific modality separately. In addition, extra auxiliary modules are injected and intertwined for cross-modal interaction. Such a design incurs an intensive computational cost that poses significant challenges to its scalability and accessibility for training and deployment. 

\begin{figure}[tb]
  \centering
  \includegraphics[width=1.\linewidth]{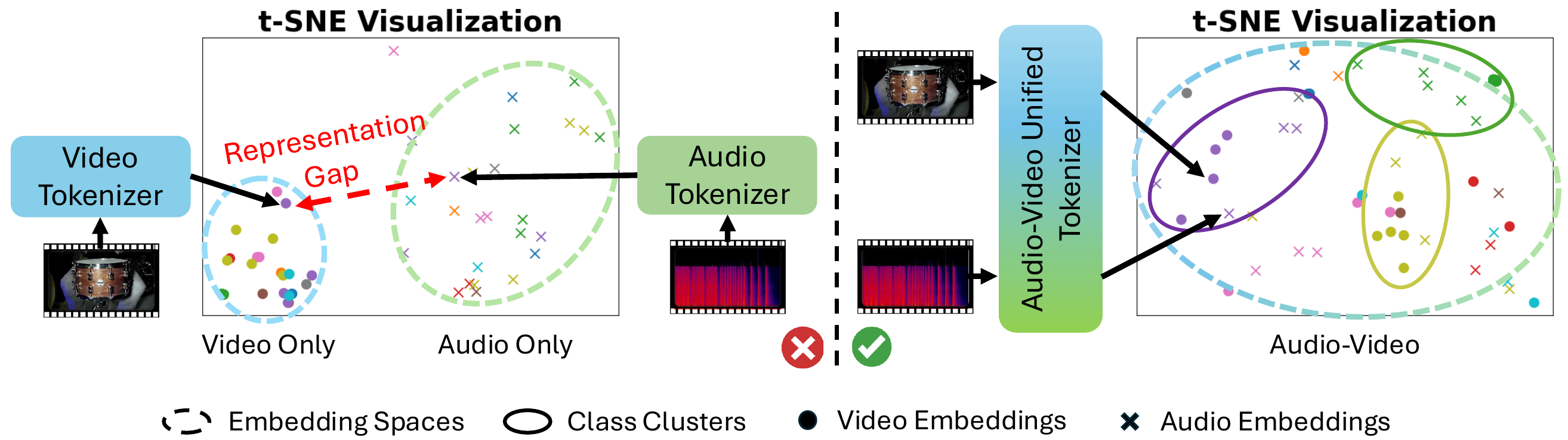}
  \caption{\textbf{Motivation.} \emph{Left:} Previous audio-video generation models typically adopt a separate pretrained tokenizer per modality and omit the representation gap between their learned embedding spaces. \emph{Right:} We aim to design a unified tokenizer that jointly encodes both modalities into a shared token space instead. Here, video and audio embeddings are colored by their respective classes.
  }
  \label{fig:motivation}
\end{figure}

Akin to single-modal predecessors~\cite{stableaudioopen, audiocraft, audiogen, ltx, cogvideox, hunyuanvideo, opensora}, various audio-video generation pipelines~\cite{ovi,javisdit,uniavgen,Ruan_2023_CVPR} tend to employ one pretrained tokenization model per modality to compress their respective input into the compact latent representation, partially alleviating the computational burden. However, such a simple integration neglects the intrinsic representation difference between the embedding spaces of the two modalities learned by those distinctively trained tokenizers, as depicted in Fig.~\ref{fig:motivation}. Inherently, their synthesized products often exhibit semantic misalignment between auditory and visual elements. To this end, an intuitive question arises: \textbf{Q:} \textit{Is it possible to jointly encode both audio and video components into a shared embedding space instead?} We hypothesize that by constructing such a shared tokenization space, not only can it avoid the mentioned representation gap to mitigate the audio-visual semantic discrepancy, but also eliminates the need to maintain an expensive dual-branch architecture for the full generation modeling. This motivates us to design a unified tokenizer for both modalities that is capable of effectively and efficiently encoding a sounding video sample into a single latent representation, holistically capturing audio-visual information for decent reconstruction and downstream AV generation tasks. 

To achieve our goal, the first critical challenge that emerges is to determine: \textbf{C1:} \textit{Which embedding representation is appropriate to unify and encapsulate auditory and visual information?} On the one hand, as raw video inputs have three-dimensional (3D) formation, the majority of prevailing video tokenizers~\cite{wfvae, vidtok, videovaeplus, opensora-plan} inherently employ 3D spatio-temporal latent as representation for compression. On the other hand, audio signals have a one-dimensional (1D) wave structure, hence many previous audio tokenizers, the so-called audio codecs~\cite{encodec, dac, unicodec, wavtokenizer}, typically encode an audio into the respective 1D temporal embedding. Meanwhile, some methods~\cite{spectralcodec, meltok} extract two-dimensional (2D) mel-spectrogram features as intermediate targets for compression and leverage neural vocoders~\cite{hifigan, bigvgan} to reconstruct raw signals more efficiently. Nevertheless, the difference in token organization (3D vs. 1D/2D) still makes it non-trivial to decide which representation is appropriate. Fortunately, some recent works~\cite{larp, adaptok} have demonstrated the potential of 1D video tokenization in constructing a causal-friendly discrete latent space that facilitates autoregressive (AR) video generation, conceptually bridging with audio's native representation. We therefore select 1D discrete latent to be the desired compact representation for audio-video encoding unification, as shown in Fig.~\ref{fig:teaser}(a).

Considering the objective 1D latent representation, the final and most important challenge to tackle is \textbf{C2:} \textit{How to design a suitable architecture to actualize 1D unified audio-video tokenization?} Based on a current state-of-the-art 1D video tokenizer~\cite{larp}, we propose \textbf{AVTok}, which is a novel attempt on this novel challenge. Drawing inspiration from~\cite{cavmae, cavmae-plus} that tackle AV pretraining tasks, we transform the baseline~\cite{larp} into a dual-stream query-based transformer with shared encoder-decoder and modal-specific queries. This model design has several characteristics: (1) Unlike~\cite{cavmae, cavmae-plus} which utilize patch-wise local-constraint information, AVTok leverages a holistic tokenization scheme with learned queries to capture higher-level, holistic AV information; (2) Dual-stream forward passes allow AVTok to harmoniously exploit auditory and visual specific elements while fusing their information implicitly to enhance reconstruction, maintaining both efficacy and efficiency; (3) AVTok inherits the AR-friendliness of the baseline~\cite{larp} that is beneficial for downstream AR-based AV generation tasks. 

Despite possessing the above-mentioned architectural advantages, training AVTok properly is challenging due to several reasons. First, visual data exhibit significantly different information density from their corresponding auditory companions, causing the model to suppress the learning and deteriorate the performance of one or both modalities. Secondly, the implicitness of information fusing via shared model parameters may lead to insufficient cross-modal interaction that hinders alignment learning. Therefore, we introduce a hierarchical training strategy: Video-First-Audio-Later (VFAL), to realize respective reconstruction capability for each individual modality in a progressive manner. Additionally, inspired by~\cite{dera, repa}, we leverage the features extracted from audio-visual foundational models~\cite{cavmae, cavmae-plus} with rich semantic correspondence to enhance model learning via a representation alignment objective. The experimental results highlighted in Fig.~\ref{fig:teaser}(b,c) show that AVTok achieves outstanding performance not only in AV reconstruction but also in downstream generation tasks, including audio-to-video, video-to-audio, and class-conditional joint AV generation.

Overall, our contributions are summarized as follows:
\begin{itemize}
    \item We propose a novel task of unified audio-video (AV) tokenization, which aims at jointly encoding both auditory and visual components into a single latent representation, facilitating efficient and effective AV reconstruction and downstream generation.
    \item We present AVTok, a 1D unified AV tokenizer attempting to fulfill the task by leveraging a multi-stream transformer-based architecture with shared encoder-decoder and modal-specific queries.
    \item We introduce VFAL, a hierarchical training paradigm equipped with a representation alignment learning objective to progressively incorporate video then audio encoding and reconstruction capabilities into AVTok.
    \item Extensive experiments highlight that AVTok excels in not only unified AV reconstruction but also downstream tasks, including audio-to-video (A2V), video-to-audio (V2A), and class-conditional joint AV generation (cJAVG).
\end{itemize}

\section{Related Work}
\label{sec:rel}

\subsection{1D Visual Tokenization}
With the Multimodal Large Language Model (MLLM) for understanding and generation tasks gaining growing popularity in recent years, 1D visual tokenization has emerged as an indispensable component. Not only does it bridge the vision-language representation gap, but it also reduces the computational burden incurred when processing visual data, enabling effortless and efficient integration of visual input into well-established LLMs. Early studies mainly focused on the image domain starting with TiTok~\cite{titok}, a transformer-based tokenizer with learnable queries that can encode a $256\times256\times3$ image using as few as 32 discrete tokens. TA-TiTok~\cite{tatitok} then uses rich semantic information from textual input to complement visual features and improve the decoding stage. Subsequent works~\cite{flextok, selftok, semanticist} enforce causality relationships among resulting tokens, making their models autoregressive (AR)-friendly for better adaptation into MLLMs. 

Recent advances have started to be explored in the video domain. LARP~\cite{larp} is the pioneer that employs a query-based transformer architecture with a holistic tokenization scheme and an autoregressive prior model to tokenize videos into a 1D latent representation with optimal token order for downstream AR generation tasks. It is then followed by Adaptok~\cite{adaptok}, which attempts to induce an adaptive temporal causality within latent space and dynamically manipulate token allocation for flexible tokenization, and DeRA~\cite{dera}, which decouples spatial-temporal representation learning to achieve more efficient and effective training. Inspired by these works and their insights, our work aims to extend the concept of 1D unified tokenization for audio and video together.

\subsection{Audio Tokenization}
Unlike image and video domains that inherently involve 2D and 3D spatial structures, audio is naturally a 1D time-varying signal representing the sound wave's amplitude over time. Audio tokenization, \textit{a.k.a} neural audio coding, has been a long-standing challenge, aiming to balance high-fidelity reconstruction with low-bitrate discrete representation that facilitates incorporation into LLMs. Some early codecs include EnCodec~\cite{encodec} and DAC~\cite{dac} that utilize residual vector quantization (RVQ) within a fully convolutional encoder-decoder architecture. Recently, UniCodec~\cite{unicodec} focuses on reducing the redundancy inherent in multi-codebook RVQ systems by constructing a unified codebook for universal sound domains. Meanwhile, SpecVQGAN~\cite{specvqgan}, Spectral Codec~\cite{spectralcodec}, and MelTok~\cite{meltok} also improve efficiency but alternatively by compressing mel-spectrograms instead of raw waveforms. With the aligned 1D discrete representation, we aim to replicate their audio tokenization capability in our unified model. 

\subsection{Audio-Video Generation}
Generative tasks involving audio and video modalities, such as audio-to-video (A2V), video-to-audio (V2A), and joint audio-video generation (JAVG) have attracted a lot of research attention in recent years, leading to a proliferation of many models with impressive synthesizing abilities. Some of the representative works for A2V generation include TempoTokens~\cite{tempotoken} that adapts a pretrained text-to-video diffusion model to support audio conditioning and achieve better synchronization, Seeing-and-Hearing~\cite{seeing-and-hearing} introduces a diffusion latent aligner to enhance cross-modal semantic coherence, and SpA2V~\cite{spa2v} harnesses spatial auditory cues to realize spatial alignment in synthesized videos.

Regarding the V2A generation, SpecVQGAN~\cite{specvqgan} is one of the early studies to train a transformer to sample spectrograms conditioning on video features from a pretrained codebook obtained by a VQGAN-variant tokenizer. Later, V-AURA~\cite{v-aura} introduces an autoregressive model with an audio-visual feature fusion strategy to enhance temporal alignment. Recently, FoleyCrafter~\cite{foleycrafter}, VinTAGe~\cite{vintage}, and MMAudio~\cite{mmaudio} leverage diffusion and flow matching generative models to achieve better audio synthesis fidelity and diversity.

By unifying the A2V and V2A goals, the JAVG task enables the joint synthesis of high-fidelity video and audio, prioritizing individual modal quality with seamless cross-modal synchronization and semantic alignment. The latest approaches~\cite{ovi, javisdit, uniavgen, avdit} primarily adopt a dual-branch architecture with separate variational autoencoder (VAE) and diffusion transformer (DiT) as tokenization and generation modules, respectively, per modality.  Despite showing impressive results, such a design is heavy-weighted and necessitates intensive computing resources for adequate training. Besides, using distinct tokenizers also neglects the representation gap between auditory and visual elements, hence they are prone to producing results with semantic misalignment. To address this problem, in this work, we introduce a unified tokenizer to jointly encode both audio and video into a single latent representation.  

\section{Method}
\label{sec:med}

\subsection{Preliminary}
\subsubsection{Query-based 1D Video Tokenization.}
\label{subsec:query1d}
As discussed in Sec.~\ref{sec:intro}, the prevailing video tokenizers predominantly adopt a 3D patch-wise tokenization scheme of which latent tokens are encoded from the 3D spatio-temporal patches of the input video, limiting them to low-level patch features and hindering the exploitation for higher-level information. To break this local constraint and enable 1D tokenization, LARP~\cite{larp} and following works~\cite{adaptok, dera} adapt the philosophy of~\cite{detr, blip2} to leverage a set of fixed learnable queries to capture holistic information in the video. Given a video input $\mathbf{V}\in\mathbb{R}^{T \times H \times W \times 3}$, it is first processed as:
$$ \mathbf{P}^v=\mathcal{P}(\mathbf{V}), \quad \mathbf{E}^v=\mathcal{F}(\mathbf{P}^v),$$
where $\mathcal{P}$ and $\mathcal{F}$ are linear patchify and flatten operations, $\mathbf{P}^v\in\mathbb{R}^{\frac{T}{f_T}\times\frac{H}{f_H}\times\frac{W}{f_W}\times d}$ and $\mathbf{E}^v\in\mathbb{R}^{m\times d}$ represent the spatiotemporal patches projected onto $d$ dimensions and their flattened embeddings. Here, $f_T, f_H, f_W$ correspond to the downsampling factors for dimensions $T, H, W$ respectively, and $m=\frac{T}{f_T}\times\frac{H}{f_H}\times\frac{W}{f_W}$ is the total number of tokens. Subsequently, a set of $n$ learnable holistic query embedding $\mathbf{Q}^v_L\in\mathbb{R}^{n\times d}$ is introduced to encode and quantize the patch embeddings $\mathbf{E}^v$ as follows:
$$\mathbf{Z}^v=\mathcal{E}(\mathbf{Q}^v_L\|\mathbf{E}^v), \quad \mathbf{x}^v=\mathcal{Q}(\mathbf{Z}^v_{1:n}),$$
in which $\mathcal{E}$ and $\mathcal{Q}$ are the encoder and quantizer, $\|$ denotes the concatenation operation, and $\mathbf{Z}^v$ is the latent embeddings of length $(n+m)$. Note that only $\mathbf{Z}^v_{1:n}$, \ie, the first $n$ ones corresponding to the query embeddings $\mathbf{Q}^v_L$ are quantized into $\mathbf{x}^v=(x^v_1,\dots,x^v_n)$ discrete tokens, ensuring each $x_v^i$ can represent any video patch equally. Eventually, during the decoding stage, another $m$ learnable patch query embeddings $\mathbf{Q}^v_P\in\mathbb{R}^{m\times d}$ are utilized to reconstruct the video as:
$$\hat{\mathbf{Z}}^v=\mathcal{Q}^{-1}(\mathbf{x}^v), \quad \hat{\mathbf{E}}^v=\mathcal{D}(\mathbf{Q}^v_P\|\hat{\mathbf{Z}}^v), \quad \hat{\mathbf{V}}=\mathcal{R}(\hat{\mathbf{E}}^v_{1:m}),$$ 
where $\mathcal{Q}^{-1}$ denotes de-quantization operation that maps discrete tokens $\mathbf{x}^v$ back to the continuous latent embedding $\hat{\mathbf{Z}}^v\in\mathbb{R}^{n\times d}$. Subsequently, they are concatenated with $\mathbf{Q}^v_P$ and go through the decoder $\mathcal{D}$ to decode $\hat{\mathbf{E}}^v$, of which only the first $m$ vectors are reshaped via the $\mathcal{R}$ operator to reconstruct $\hat{\mathbf{V}}\in\mathbb{R}^{T \times H \times W \times 3}$.

\begin{figure}[tb]
  \centering
  \includegraphics[width=1.\linewidth]{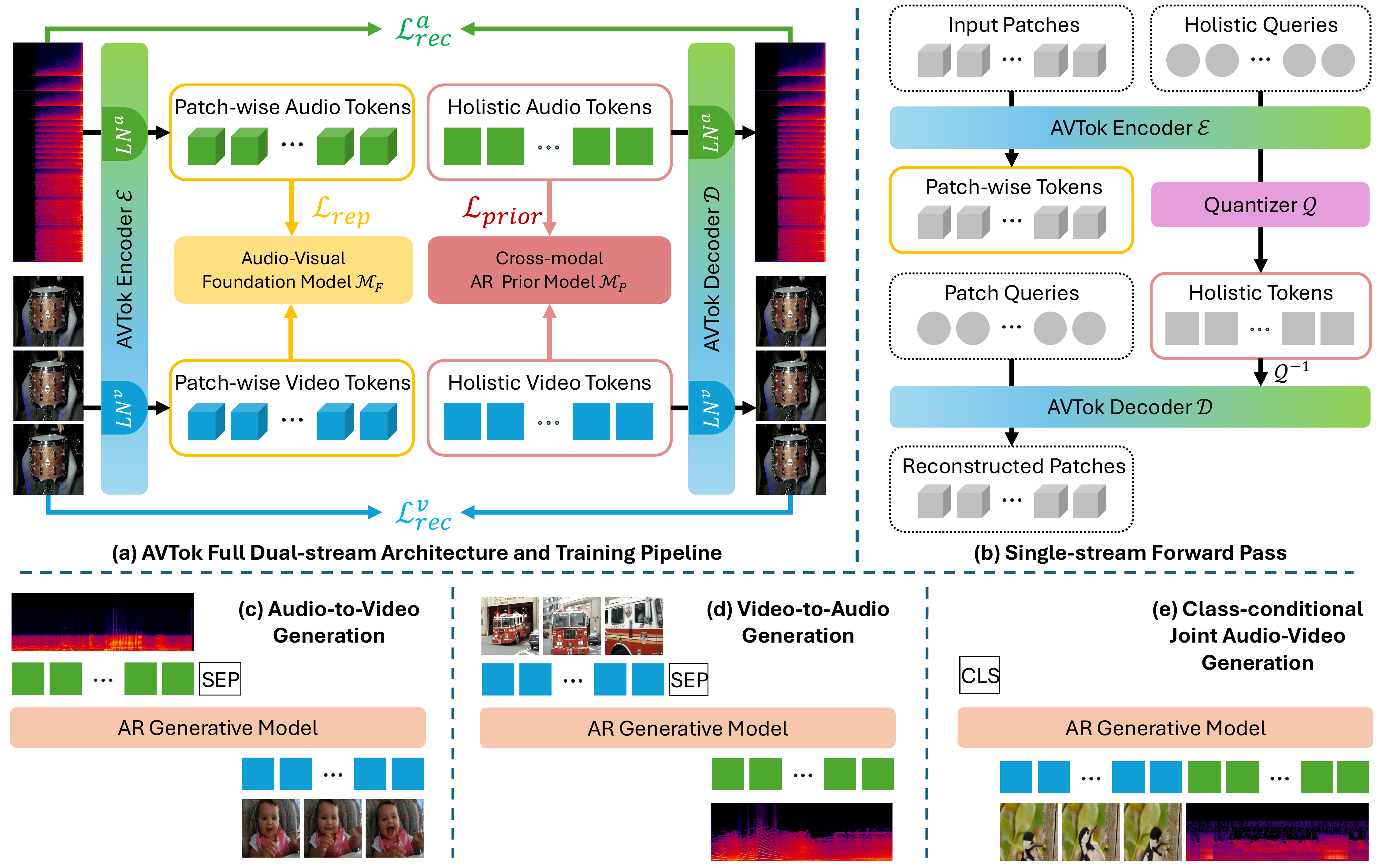}
  \caption{\textbf{Method illustration.} Cubes \manimpossiblecube, squares $\square$, and circles $\Circle$ respectively represent input patches or patch-wise tokens, holistic discrete tokens, and continuous query embeddings. \textbf{(a) AVTok} features a dual-stream transformer-based architecture, of which each stream's forward pass is demonstrated in \textbf{(b)}, to jointly learns video ({\color{cyan}Blue} stream) and audio ({\color{green}Green} stream) reconstructions in a unified holistic scheme. It leverages separate sets of learnable queries and normalization layers to gather modal-specific information, while \textit{sharing} remaining parameters to enable implicit cross-modal interaction, achieving both efficiency and efficacy. In addition to the standard reconstruction training objectives {\color{cyan}$\mathcal{L}_{rec}^v$} and {\color{green}$\mathcal{L}_{rec}^a$}, we align AVTok's patch-wise continuous tokens with an audio-visual foundation model $\mathcal{M}_F$ via {\color[HTML]{FFDF00}$\mathcal{L}_{rep}$} to better capture synergistic features between auditory and visual elements. Lastly, an AR prior model $\mathcal{M}_P$ is also equipped to encourage an AR-friendly discrete latent space via {\color{purple}$\mathcal{L}_{prior}$}, facilitating downstream AR generative tasks including \textbf{(c) audio-to-video}, \textbf{(d) video-to-audio}, and \textbf{(e) class-conditional joint audio-video} generation.}
  \label{fig:pipeline}
\end{figure}

\subsubsection{Autoregressive Generative Prior.} Although the 1D latent tokens $\mathbf{x}^v$ obtained with the aforementioned query-based tokenizer are now holistic and discrete, there is no specific flattening order enforced. This is because of the unordered nature of the holistic query set and the parallel processing property of the transformer encoder. To make such a latent space compatible with AR generative models, LARP~\cite{larp} incorporates a lightweight AR transformer with adjusted input and output layers as prior model $\mathcal{M}_P$ to provide gradients for structure optimization. It is jointly trained with the tokenizer in an end-to-end manner using negative log-likelihood (NLL) loss $\mathcal{L}_{prior}$ for next token prediction objective (NTP) in synergy with reconstruction loss $\mathcal{L}^v_{rec}$ as:
$$\mathcal{L}=\mathcal{L}^v_{rec}+\alpha\mathcal{L}_{prior},$$
where $\alpha$ is the loss weight. Notably, this prior model serves the sole purpose of promoting an AR-friendly discrete latent space during training. It is discarded during inference and thus affects neither the speed nor the memory footprint.

\subsection{Holistic Audio-Video Tokenization}
\label{subsec:3.2}
\subsubsection{Patchify.} AVTok employs the architecture described in Sec.~\ref{subsec:query1d} as its video stream of which the patchification remains unchanged that transforms a video input $\mathbf{V}\in\mathbb{R}^{T \times H \times W \times 3}$ into a flattened d-dimensional embedding $E^v\in\mathbb{R}^{m\times d}$. For audio stream, instead of $\mathbf{A}_{raw}\in\mathbb{R}^{N}$ which is 1D-long continuous data, we opt to use its normalized mel-spectrogram $\mathbf{A}_{mel}\in\mathbb{R}^{M\times L}$ as input. Here, $M$ and $L$ denote the number of frequency bins and time frames, respectively. Not only does $\mathbf{A}_{mel}$ reduce computation complexity, but it can also be interpreted as a gray-scale image that can be patchified similarly as in video stream. Notably, it can be converted back to the raw waveform with lossless quality using off-the-shelf vocoders~\cite{hifigan,bigvgan}. Given $\mathbf{A}_{mel}$, herein referred to as $\mathbf{A}$ for brevity, we process it as: 
$$\mathbf{P}^a=\mathcal{P}(\mathbf{A}), \quad \mathbf{E}^a=\mathcal{F}(\mathbf{P}^a),$$
where $\mathbf{P}^a\in\mathbb{R}^{\frac{M}{f_M}\times\frac{L}{f_L}\times d}$ and $\mathbf{E}^a\in\mathbb{R}^{p\times d}$ represent the audio patches projected onto $d$ dimensions and their flattened embeddings. Here, $f_M,f_L$ correspond to downsampling factors for dimension $M, L$ accordingly, and $p=\frac{M}{f_M}\times\frac{L}{f_{L}}$ is the total number of audio tokens. 

\subsubsection{Dual-stream Transformer.} Although transformer-based design has been applied to the context of 1D tokenization for both audio and video modality individually referring to Sec.~\ref{sec:rel}, it has never been explored for the unified setting involving both modalities simultaneously. As the initial attempt for this work, we extend the query-based design in~\cite{larp} from the video-only modality to audio-video multi-modality and build a single-stream \textit{vanilla} version of our AVTok tokenizer. Given the patchified $\mathbf{E}^a$ and $\mathbf{E}^v$ embeddings obtained above, we concatenate them and construct a joint embedding $\mathbf{E}^{av}=(\mathbf{E}^v \| \mathbf{E}^a)\in\mathbb{R}^{(m+p)\times d}$. It will then be encoded, quantized, and decoded similarly following Sec.~\ref{subsec:query1d} as:
$$\mathbf{x}^{av}=\mathcal{Q}(\mathcal{E}(\mathbf{Q}^{av}_L\|\mathbf{E}^{av})_{1:n}), \quad \hat{\mathbf{E}}^{av}=\mathcal{D}(\mathbf{Q}_P^{av}\|\mathcal{Q}^{-1}(\mathbf{x}^{av})), $$
$$\hat{\mathbf{V}}=\mathcal{R}(\hat{\mathbf{E}}^{av}_{1:m}), \quad \hat{\mathbf{A}}=\mathcal{R}(\hat{\mathbf{E}}^{av}_{m:m+p}),$$
where $\mathbf{Q}_L^{av}\in\mathbb{R}^{n\times d}$ and $\mathbf{Q}_P^{av}\in\mathbb{R}^{(m+p)\times d}$ denotes learnable holistic and patch query embeddings respectively. This simple design features cross-modal modeling that may help the model to exploit audio-visual correlation to reconstruct one modality based on the information of the other. However, without explicitly considering the modal-specific features, their significant difference in nature often causes the \textit{vanilla} model to train inadequately in which the learning of one modality harms that of the other, eventually yielding subpar performance.

To alleviate this problem, we adapt the philosophy of~\cite{cavmae, cavmae-plus} to bootstrap the \textit{vanilla} design into a dual-stream architecture with shared encoder-decoder but separate sets of learnable holistic and patch query embeddings as well as normalization layers for our finalized AVTok tokenizer. Specifically, we input audio and video patch embeddings $\mathbf{E}^a$ and $\mathbf{E}^v$ in two different forward passes to the encoder $\mathcal{E}(\cdot;LN_1,LN_2)$ then decoder $\mathcal{D}(\cdot;LN_1,LN_2)$ with each stream leveraging a separate set of normalization layers $\big(LN_1^{\{a,v\}}, LN_2^{\{a,v\}}\big)$ as follows:
\begin{equation*}
    \begin{gathered}
    \mathbf{Z}^i=\mathcal{E}(\mathbf{Q}^{i}_L\|\mathbf{E}^{i}; LN_1^i, LN_2^i), \enspace \mathbf{x}^{i}=\mathcal{Q}(\mathbf{Z}^i_{1:j}), \enspace \hat{\mathbf{E}}^{i}=\mathcal{D}(\mathbf{Q}_P^{i}\|\mathcal{Q}^{-1}(\mathbf{x}^{i}); LN_1^i, LN_2^i), \\
    \hat{\mathbf{V}}=\mathcal{R}(\hat{\mathbf{E}}^{v}_{1:m}), \quad \hat{\mathbf{A}}=\mathcal{R}(\hat{\mathbf{E}}^{a}_{1:p}), \quad (i, j)\in\{(v, n), (a, q)\},
    \end{gathered}
\end{equation*}
where $\mathbf{Q}_L^{v}\in\mathbb{R}^{n\times d}, \mathbf{Q}_P^{v}\in\mathbb{R}^{m\times d}, \mathbf{Q}_L^{a}\in\mathbb{R}^{q\times d}, \mathbf{Q}_P^{a}\in\mathbb{R}^{p\times d}$ respectively represent the learnable holistic and patch query embeddings of video and audio modality. This design facilitates harnessing modal-specific information by using distinctive learnable components per modality, while still allowing for implicit audio-visual fusion via sharing remaining parameters, thereby achieving both efficiency and effectiveness for reconstruction and downstream generation tasks. The detailed illustration of AVTok is shown in Fig.~\ref{fig:pipeline}(a, b).
\subsubsection{Reconstruction Objective.} Following the composition in~\cite{larp}, the reconstructive training loss for the video stream of AVTok, \ie $\mathcal{L}_{rec}^v$, is constituted by $L_1$ reconstruction loss, LPIPS perceptual
loss~\cite{lpips}, GAN adversarial loss~\cite{gan}, and SVQ quantization loss~\cite{larp}. Meanwhile, for $\mathcal{L}_{rec}^a$, since the audio stream reconstruction process involves pretrained vocoders~\cite{hifigan, bigvgan}, we follow them to adopt Multi-Scale Mel-Spectrogram Loss~\cite{dac} as reconstruction loss, use Multi-Scale Sub-Band CQT Discriminator~\cite{cqt} and Multi-Period Discriminator~\cite{hifigan} for adversarial components, and reuse SVQ quantization loss from the video stream.

\subsection{Hierarchical Training Paradigm} 

\subsubsection{Video-First-Audio-Later (VFAL) Strategy.} Despite having several architectural advantages, our experiments reveal that simply training AVTok from scratch is non-ideal. This is primarily because of the fact that visual information is abundant, which dominates auditory information, causing the learning of the video stream to suppress that of the audio stream. To accommodate this issue, we design the VFAL hierarchical training strategy for optimal and efficient training of AVTok. Specifically, we start with the training of the more challenging modality, \ie, video stream, while discarding the audio stream in Stage 1, aiming to realize reconstruction ability for visual elements and establish a strong latent token representation space. Subsequently, in Stage 2, we reattach and train only the modules specialized for the audio stream while freezing those of the video stream together with the shared ones, realizing audio reconstruction capability. This is intuitively possible considering that the input mel-spectrogram can be treated as a gray-scale image, as mentioned in Sec.~\ref{subsec:3.2}. Finally, in the last stage, we finetune the decoding modules to attain unified audio-video reconstruction with refined quality. By imposing this explicit training path, VFAL encourages AVTok to optimize the learning of each stream progressively.

\subsubsection{Representation Alignment Learning.} During experiments, we also observed an issue in which AVTok does not fully exploit audio-visual correspondent features to improve the final reconstruction. We hypothesize that this might be because the cross-modal interaction via shared model parameters is implicit, and hence it hinders audio-visual alignment learning. Drawing inspiration from~\cite{dera, repa}, we
leverage a pretrained audio-visual foundation model $\mathcal{M}_F$~\cite{cavmae-plus} that learned an embedding space with rich semantics and strong correspondence between visual and auditory information as the intermediate aligning module to enhance cross-modal alignment between the two streams of AVTok. This can be achieved by incorporating into the training the representation alignment objective $\mathcal{L}_{rep}$, which can be computed as follows:
\begin{equation*}
    \begin{gathered}
    \mathbf{Z}^v_F=\mathcal{M}_F(\mathbf{V}), \quad \mathbf{Z}^a_F=\mathcal{M}_F(\mathbf{A}), \quad \tilde{\mathbf{Z}}^v=\mathbf{Z}^v_{n:m+n}, \quad \tilde{\mathbf{Z}}^a=\mathbf{Z}^a_{q:p+q}, \\
    \mathcal{L}_{rep}=-\mathbb{E}\Big[\sum_{i\in\{a,v\}}\frac{1}{N_i}\sum_{k=1}^{N_i}\text{sim}(\mathbf{Z}_F^{i}[k], h_\phi(\text{interp}(\tilde{\mathbf{Z}}^i)[k]))\Big],
    \end{gathered}
\end{equation*}
where $\mathbf{Z}^v_F, \mathbf{Z}^a_F, \tilde{\mathbf{Z}}^v, \tilde{\mathbf{Z}}^a$ denote the video and audio patch embeddings of length $N_v,N_a,m,p$ extracted by $\mathcal{M}_F$ and our AVTok's encoder, $k$ is a patch index, $\text{sim}(\cdot, \cdot)$ is a pre-defined similarity function, and $h_\phi$ represents a multilayer perceptron (MLP). Similarly to~\cite{dera, repa}, we linearly interpolate $\tilde{\mathbf{Z}}^v, \tilde{\mathbf{Z}}^a$ to the same length of $\mathbf{Z}^v_F, \mathbf{Z}^a_F$ via $\text{interp}(\cdot)$ operator for computational compatibility. 
\subsubsection{Cross-modal AR Generative Prior.} To facilitate the downstream audio-to-video, video-to-audio generation, and class-conditional joint audio-video generation tasks simultaneously, 
we adapt the autoregressive generative prior of~\cite{larp} mentioned in Sec.~\ref{subsec:query1d} by simply computing the NTP objective loss for two token orders $\mathbf{x}^v\|\mathbf{x}^a$ and $\mathbf{x}^a\|\mathbf{x}^v$ to compose $\mathcal{L}_{prior}$. Finally, the overall training objective for AVTok is formulated as:
$$\mathcal{L}=\lambda_1\mathcal{L}_{rec}^v+\lambda_2\mathcal{L}_{rec}^a+\lambda_3\mathcal{L}_{rep}+\lambda_4\mathcal{L}_{prior},$$
with $\lambda_{1,2,3,4}$ are the loss weights for each component. 

\section{Experiments}
\label{sec:exp}
\begin{table}[tb]
  \caption{\textbf{Quantitative comparison of reconstruction results.} Results are categorized into video-only (VO), audio-only (AO), and joint audio-video (AV) tokenization functionality. W/M denote resolution of waveform/mel-spectrogram used as audio input. The best and second best results are in \textbf{bold} and \underline{underlined}.
  }
  \label{tab:tokenization}
  \centering
  \renewcommand{\arraystretch}{1.25}
    \resizebox{1.0\textwidth}{!}{
        \begin{tabular}{c|r|p{21mm}<{\centering} p{14mm}<{\centering} |p{14mm}<{\centering} p{14mm}<{\centering} p{14mm}<{\centering}|p{14mm}<{\centering} p{14mm}<{\centering} p{17mm}<{\centering}}
            \Xhline{2.0\arrayrulewidth}
            \multirow{2}{*}{Type} & \multirow{2}{*}{Method} & \multicolumn{2}{c|}{Configuration} & \multicolumn{3}{c|}{Video Metrics} &\multicolumn{3}{c}{Audio Metrics}\\
            & & Resolution & \#Tokens & PSNR$\uparrow$ & rFVD$\downarrow$ & LPIPS$\downarrow$ & SI-SDR$\downarrow$ & rFAD$\downarrow$ & MR-STFT$\downarrow$ \\
            \Xhline{2.0\arrayrulewidth}
            \multirow{3}{*}{VO} & OmniTokenizer~\cite{omnitokenizer} & $17\times128 \times128$ & 1280 & 23.84 & 90.99 & 0.203 & - & - & - \\
             & AdapTok~\cite{adaptok} & $16\times128\times128$ & 2048 & 23.87 & 22.23 & 0.180 & - & - & - \\
             & LARP~\cite{larp} & $16\times128\times128$ & 1024 & 24.53 & 14.24 & 0.137 & - & - & - \\
            \Xhline{2.0\arrayrulewidth}
            \multirow{3}{*}{AO} & WavTokenizer~\cite{wavtokenizer} & $98304\times1$ (W) & 164 & - & - & - & 24.27 & 6.82 & 1.589 \\
            & UniCodec~\cite{unicodec} & $98304\times1$ (W) & 308 & - & - & - & \textbf{18.25} & 6.73 & \textbf{1.508} \\
            & SpectralCodec~\cite{spectralcodec} & $80\times384$ (M) & 384 & - & - & - & 29.30 & \textbf{5.56} & 1.514 \\
            \Xhline{2.0\arrayrulewidth}
             \rowcolor{gray!10} AV & \textit{Vanilla} & $16\times128\times128$ & & 24.50 & 14.87 & 0.140 & 35.45 & 10.26 & 2.114 \\
            \rowcolor{gray!10} (Ours) & AVTok & $80\times384$ (M) & \multirow{-2}{*}{1152} & \textbf{25.62} & \textbf{12.80} & \textbf{0.126} & \underline{23.09}& \underline{5.93} & 1.523\\
            \Xhline{2.0\arrayrulewidth}
        \end{tabular}
    }
\end{table}
\begin{figure}[tb]
  \centering
  \includegraphics[width=1.\linewidth]{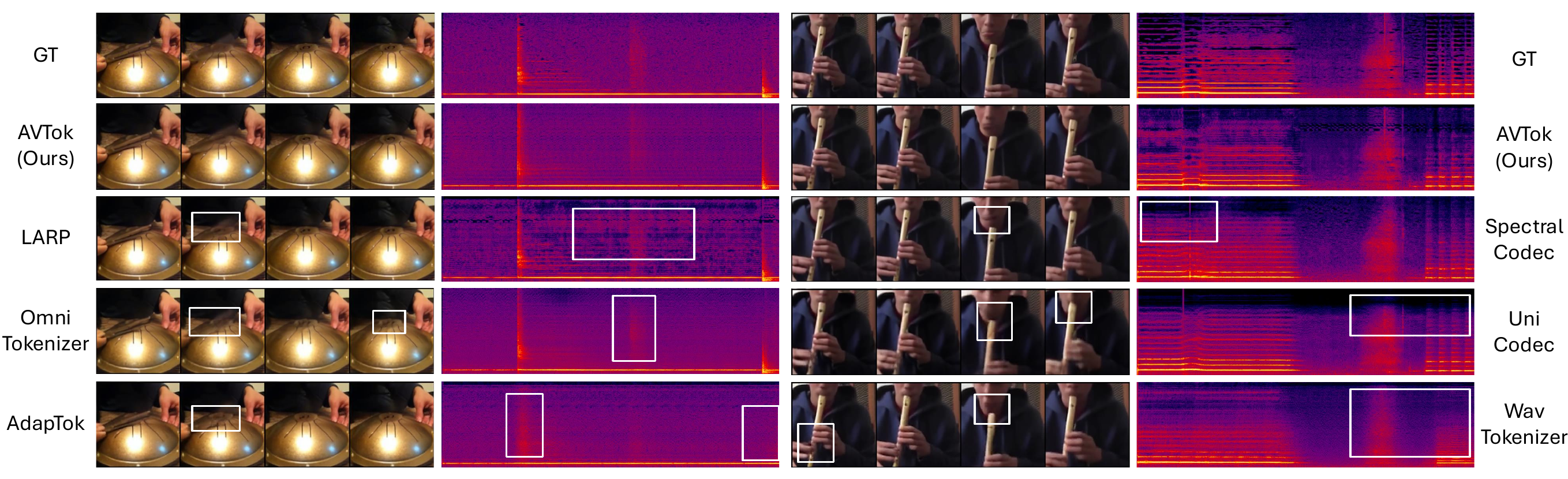}
  \caption{\textbf{Qualitative comparison of reconstruction results.}
  }
  \label{fig:rec_qualitative}
\end{figure}
\subsection{Setup}
\subsubsection{Dataset.} We conduct our experiments on TAVGBench~\cite{tavgbench} and VGGSound~\cite{vggsound} datasets. Both are used in the reconstruction whilst only the latter is used in the downstream generation tasks deliberately for demonstration purposes due to time and resource constraints. The test set of VGGSound is used for all assessments. By default, we use 16-frame sounding video clips with spatial resolution $128\times128$, frame rate of $3.6$fps, single-channel audio with waveform resolution $98304\times1$, $22$kHz sampling rate, and mel-spectrogram resolution $80\times384$ in both training and evaluation.   

\subsubsection{Implementation Details.} For patchification, we first follow~\cite{larp,ast} to split the input video and audio mel-spectrogram into continuous visual and auditory patch embeddings using $(f_T, f_H, f_W)=(4, 8, 8)$ and $(f_M, f_L)=(16, 16)$ respectively. We then utilize a set of $n=1024, q=128$ learnable holistic queries to obtain $1152$ holistic discrete tokens. For decoding, another set of $m=1024, q=120$ learnable patch queries are leveraged to reconstruct their corresponding modality. Besides, we use HiFi-GAN~\cite{hifigan} to convert the output mel-spectrograms back to waveforms. For the remaining, unless otherwise specified, we maintain the same configuration as~\cite{larp} by default. Regarding downstream generation tasks, we also follow~\cite{larp} to adopt Llama-like transformer~\cite{llama,llamagen} to be our AR generative model. As shown in Fig.~\ref{fig:pipeline}(c-e), one class token \texttt{[CLS]} is used in the class-conditional joint audio-video generation task, while one separator token \texttt{[SEP]} is employed for cross-modal generation tasks.
\begin{table}[tb]
  \caption{\textbf{Comparison of generation results.} Results are grouped by tasks including audio-to-video (A2V), video-to-audio (V2A), and class-conditional joint audio-video generation (cJAVG). Diff, AR, and FM respectively denote diffusion, autoregressive, and flow matching generative paradigms. The best and second best results are in \textbf{bold} and \underline{underlined}.
  }
  \label{tab:generation}
  \centering
  \renewcommand{\arraystretch}{1.25}
    \resizebox{1.\textwidth}{!}{
        \begin{tabular}{c|r|c|c c|p{14mm}<{\centering} p{14mm}<{\centering} p{14mm}<{\centering} p{14mm}<{\centering}}
            \Xhline{2.0\arrayrulewidth}
            \multirow{2}{*}{Task}& \multirow{2}{*}{Method}& Gen& \multicolumn{2}{c|}{\#Param}& \multirow{2}{*}{gFVD$\downarrow$} & \multirow{2}{*}{gFAD$\downarrow$} & \multirow{2}{*}{DeSync$\downarrow$} & \multirow{2}{*}{IB-Score$\uparrow$} \\
            & & Type & Tokenizer & Generator & & & \\
            \Xhline{2.0\arrayrulewidth}
             & TempoTokens~\cite{tempotoken} & Diff & 83.7M & 1.9B & 786.61 & - & 1.359 & 0.132 \\
             \multirow{-2}{*}{A2V} & \cellcolor{gray!10}{AVTok-A2V (Ours)} & \cellcolor{gray!10}{AR} & \cellcolor{gray!10}{208.4M} & \cellcolor{gray!10}{632.0M} & \cellcolor{gray!10}{\textbf{150.26}} & \cellcolor{gray!10}{-} & \cellcolor{gray!10}{\textbf{1.317}} & \cellcolor{gray!10}{\textbf{0.143}}  \\
            \Xhline{2.0\arrayrulewidth}
            \multirow{5}{*}{V2A} & MMAudio~\cite{mmaudio} & FM & 298.5M & 1.3B & - & \textbf{17.09} & \textbf{0.813} & \textbf{0.291}\\
            & VinTAGe~\cite{vintage} & FM & 110.6M & 1.5B &  - & 80.06 & 1.294 & 0.044\\
            & V-AURA~\cite{v-aura} & AR & 76.7M & 816.9M & - & 126.92 & \underline{0.967} & 0.231 \\
            & SpecVQGAN~\cite{specvqgan} & AR & 76.4M & 332.4M & - & 210.07 & 1.291 & 0.100 \\
            & \cellcolor{gray!10}{AVTok-V2A (Ours)} & \cellcolor{gray!10}{AR} & \cellcolor{gray!10}{208.4M} & \cellcolor{gray!10}{632.0M} & \cellcolor{gray!10}{-} & \cellcolor{gray!10}{\underline{49.47}} & \cellcolor{gray!10}{1.239} & \cellcolor{gray!10}{\underline{0.249}} \\
            \Xhline{2.0\arrayrulewidth}
            \multirow{3}{*}{cJAVG} & JavisDiT~\cite{javisdit} & FM & 448.7M & 8.9B & 1040.28 & 268.51 & 1.330 & \underline{0.195}\\
            & Ovi~\cite{ovi} & FM & 988.6M & 17.3B & \underline{972.65} & \underline{129.02} & \textbf{0.814} & 0.172 \\ 
            & \cellcolor{gray!10}{AVTok-cJAVG (Ours)} & \cellcolor{gray!10}{AR} & \cellcolor{gray!10}{208.4M} & \cellcolor{gray!10}{632.4M} & \cellcolor{gray!10}{\textbf{138.80}} & \cellcolor{gray!10}{\textbf{56.58}} & \cellcolor{gray!10}{\underline{1.319}} & \cellcolor{gray!10}{\textbf{0.206}} \\
            \Xhline{2.0\arrayrulewidth}
        \end{tabular}
    }
\end{table}
\begin{figure}[tb]
  \centering
  \includegraphics[width=1.\linewidth]{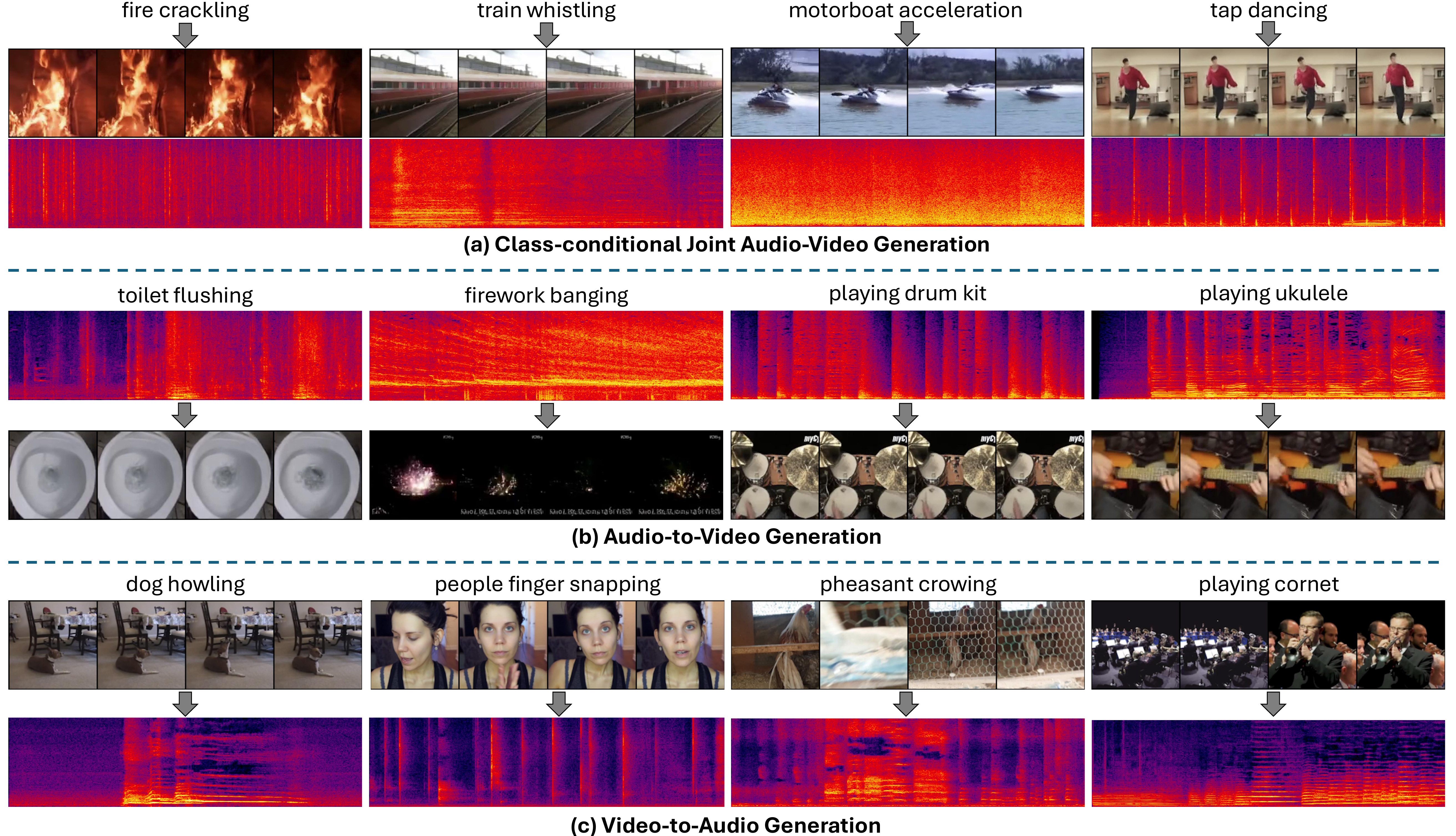}
  \caption{\textbf{Qualitative results for downstream generation tasks.} Note that class conditions are only inputted for 
  \textbf{(a)} class-conditional joint audio-video generation, and displayed here for illustration purposes of \textbf{(b)} audio-to-video and \textbf{(c)} video-to-audio generation.
  }
  \label{fig:gen_qualitative}
\end{figure}
\subsection{Reconstruction Evaluation}
\subsubsection{Baselines \& Metrics.} 
Due to the novelty of the unified audio-video tokenization task, there is no open-source baseline available for direct comparison. Therefore, in addition to our \textit{vanilla} model, we select some state-of-the-art unimodal methods from each side as representatives for comparisons including OmniTokenizer~\cite{omnitokenizer}, AdapTok~\cite{adaptok}, LARP~\cite{larp} as video-only baselines, and WavTokenizer~\cite{wavtokenizer}, UniCodec~\cite{unicodec}, SpectralCodec~\cite{spectralcodec} as audio-only baselines. Regarding metrics, we adopt PSNR~\cite{psnr}, FVD~\cite{fvd}, LPIPS~\cite{lpips} to assess video reconstruction, and employ SI-SDR~\cite{sisdr}, FAD~\cite{fad}, MR-STFT~\cite{mrstft} to evaluate audio side.

\subsubsection{Main Results.} As shown in Tab.~\ref{tab:tokenization} and Fig.~\ref{fig:rec_qualitative}, AVTok consistently outperforms the \textit{vanilla} design and selected unimodal baselines in video reconstruction while maintaining competitive performance on the audio side. They not only indicate the feasibility of the unified audio-video tokenization task and the effectiveness of our approach but also infer that leveraging cross-modal information can boost the performance of each modality.
\subsection{Generation Evaluation}
\subsubsection{Baselines \& Metrics.} For downstream generation tasks, we select and compare our AR generative models with several representative baselines, including: (1) TempoTokens~\cite{tempotoken} for audio-to-video generation (A2V); (2) MMAudio~\cite{mmaudio}, VinTAGe~\cite{vintage}, V-AURA~\cite{v-aura}, SpecVQGAN~\cite{specvqgan} for video-to-audio generation (V2A); and (3) JavisDiT~\cite{javisdit}, Ovi~\cite{ovi} for class-conditional joint audio-video generation (cJAVG). Since some of them require textual caption as the condition to control the synthesis, we bypass it by utilizing class labels as an alternative. Regarding metrics, we use FVD~\cite{fvd} and FAD~\cite{fad} to assess the quality of the video and audio samples generated, DeSync~\cite{syncformer} to measure their temporal synchronization, and ImageBind~\cite{imagebind} (IB) Score to evaluate semantic alignment.

\subsubsection{Main Results.} As demonstrated in Tab.~\ref{tab:generation}, our AR generative models incorporated with the proposed AVTok tokenizer achieve outstanding results in downstream tasks that surpass the majority of selected baselines whilst having efficient designs. This can be attributed to the learned unified discrete latent space of AVTok illustrated in Fig.~\ref{fig:motivation} which is suitable for AR audio-video generation, enabling the synthesis of high-fidelity samples as displayed in Fig.~\ref{fig:gen_qualitative}.  

\subsection{Ablation Study}
To evaluate the impact of architecture design and each training component proposed in Sec.~\ref{sec:med}, we conduct an ablation study, of which the results are shown in Tab.~\ref{tab:abla}. 

\subsubsection{Architecture Design.} We first observe that AVTok's dual-stream architecture attains superior performance compared to the single-stream \textit{vanilla} version across all tasks. This can be attributed to its capability to harness modal-specific information via distinctive learnable queries per modality while allowing for implicit cross-modal interaction via remaining shared parameters. Notably, such a design of AVTok facilitates effortless integration into AR generative models for different downstream generation tasks whereas the \textit{vanilla} one does not. 

\subsubsection{Training Components.} It is demonstrated that the hierarchical VFAL strategy and the representation alignment training objective $\mathcal{L}_{rep}$ contribute significantly to the final performance in reconstruction, which eventually benefits downstream generation. This highlights their effectiveness in encouraging AVTok to optimize the learning of each stream progressively and enhance their semantic alignment. Finally, similar to~\cite{larp}, we find that ablating cross-modal AR generative prior $\mathcal{L}_{prior}$ yields the best reconstruction but the worst synthesis results, further validating the advantage of leveraging AR prior to train an AR-friendly tokenizer for downstream generation tasks.  

\begin{table}[tb]
  \caption{\textbf{Ablation study on the impact of each component.}
  }
  \label{tab:abla}
  \centering
  \renewcommand{\arraystretch}{1.25}
    \resizebox{0.8\textwidth}{!}{
        \begin{tabular}{l|p{14mm}<{\centering} p{14mm}<{\centering}|p{14mm}<{\centering} p{14mm}<{\centering} p{14mm}<{\centering} p{14mm}<{\centering}}
            \Xhline{2.0\arrayrulewidth}
             \multirow{2}{*}{Configuration} & \multicolumn{2}{c|}{AV Reconstruction} & A2V & V2A & \multicolumn{2}{c}{cJAVG}\\
             & rFVD$\downarrow$ & rFAD$\downarrow$ &{gFVD$\downarrow$} & {gFAD$\downarrow$} & {gFVD$\downarrow$} & {gFAD$\downarrow$} \\
            \Xhline{2.0\arrayrulewidth}
             Vanilla & 14.87 & 10.26 & - & - & - & - \\
             \rowcolor{gray!10} AVTok & \underline{12.80} & \underline{5.93} & \textbf{150.26} & \textbf{49.47} & \textbf{138.80} & \textbf{56.58} \\
             \Xhline{2.0\arrayrulewidth}
             Without VFAL & 13.19 & 9.38 & 209.33 & 61.02 & 193.28 & 80.78\\
             Without $\mathcal{L}_{rep}$ & 12.90 & 8.48 & 182.15 & 54.16 & 184.20 & 75.09 \\
             Without $\mathcal{L}_{prior}$ & \textbf{10.63} & \textbf{3.47} & 266.82 & 67.84 & 249.47 & 90.11\\
            \Xhline{2.0\arrayrulewidth}
        \end{tabular}
    }
\end{table}
\section{Conclusion}
\label{sec:concl}
We have presented \textbf{AVTok}, a novel unified audio-video tokenizer capable of jointly encoding an audio-video pair into a single compact one-dimensional latent representation with a unified codebook. AVTok features a dual-stream transformer-based architecture with shared encoder-decoder and modal-specific learnable holistic queries to harmoniously exploit auditory and visual specific elements while fusing their information implicitly for efficient and effective reconstruction. To train AVTok properly, we devise Video-First-Audio-Later (VFAL), a hierarchical strategy that encourages the model to progressively develop reconstruction capability for each individual modality. Additionally, we incorporate an audio-visual foundation model to enhance cross-modal correspondence learning of AVTok via representation alignment loss, eventually improving the learning of each stream. The experimental results demonstrate not only the feasibility of the proposed unified tokenization goal but also the superiority of our model in both reconstruction and downstream generation tasks. We hope that this work will encourage further exploration in this direction to build unified large multimodal models for audio-video generation in future.
%
%
\subsubsection*{\ackname} This work was supported by National Natural Science Foundation of China (NSFC) Young Scientists Fund Category B (62522216), National Natural Science Foundation of China (NSFC) Young Scientists Fund Category C (62402408), Hong Kong SAR Research Grants Council (RGC) Early Career Scheme (26208924), Hong Kong SAR Research Grants Council (RGC) General Research Fund (16219025), and HKUST (WEB25EG01).

\bibliographystyle{splncs04}
\bibliography{main}

\newpage
\appendix

\section{Additional Experiment Details}

\subsection{Datasets}
\subsubsection{Statistics.} We conduct our experiments on VGGSound~\cite{vggsound} and TAVGBench~\cite{tavgbench} datasets. VGGSound consists of more than 210K sounding video clips spanning across 310 different classes and is commonly used in various audio-visual understanding and generation tasks. Due to data corruption, only approximately 200K audio-video pairs are available for our usage, of which the train split contains 180K samples and the remaining samples belong to the test split. Meanwhile, TAVGBench is a larger-scale dataset containing 1.7M samples with better alignment between auditory and visual elements compared to VGGSound. However, we only utilize a subset of 460K high-quality samples filtered by~\cite{javisdit} through a series of filtering strategies to accommodate resource constraints.

\subsubsection{Composition.} Eventually, a total of 640K data from both TAVGBench and the train split of VGGSound are used to train the AVTok tokenizer for the reconstruction task, while only the 180K VGGSound ones are used to train AR generative models for the downstream generation tasks. The test set of VGGSound is used for all evaluations. All audio-video pairs are preprocessed following the adopted neural vocoder HiFi-GAN~\cite{hifigan} and the baseline 1D video tokenizer LARP~\cite{larp}, respectively, to the default input resolutions mentioned in the main text, while ensuring that they are synchronized in the temporal dimension with the duration deliberately set at around 4 seconds. This facilitates that the audio component is long enough to provide sufficient and meaningful auditory information for model training while maintaining efficiency and compatibility with the adopted pretrained audio-visual foundation model~\cite{cavmae-plus}.

\subsection{Model Implementation}

\subsubsection{AVTok Tokenizer.} We follow LARP~\cite{larp} to adopt fixed sin-cos positional encoding~\cite{attention} in both the encoder and decoder of AVTok. In the encoder, fixed 3D and 2D positional encodings are applied to each video and audio patch, while in the decoder, fixed 1D positional encodings are added to each holistic video and audio token. Notably, since the patch queries and holistic queries for both modalities are position-wise learnable parameters, they do not necessitate additional positional encodings.

The encoder and decoder of AVTok adopt the standard transformer design~\cite{attention} in which each layer consists of multi-headed self-attention ($MSA$), layer normalization ($LN_1, LN_2$), and multilayer perceptron ($MLP$) blocks with residual connections. Specifically, the forward pass of each layer is as follows:
\begin{equation*}
    \begin{gathered}
    \mathbf{x'}=MSA(LN_1(\mathbf{x})) + \mathbf{x}, \quad \mathbf{y}=MLP(LN_2(\mathbf{x'})) + \mathbf{x'},
    \end{gathered}
\end{equation*}
where $\mathbf{x}$ represents the concatenation of learnable holistic queries/patch queries with input patches/holistic tokens described in the main text. We then adapt the philosophy of~\cite{cavmae,cavmae-plus} to use separate sets of $(LN_1^a, LN_2^a)$ and $(LN_1^v, LN_2^v)$ for audio and video streams to efficiently formulate the final dual-stream architecture.

We employ HiFi-GAN~\cite{hifigan}, CAV-MAE Sync~\cite{cavmae-plus}, and GPT-2~\cite{gpt2} to be our neural vocoder, audio-visual foundational model $\mathcal{M}_F$, and cross-modal AR generative prior model $\mathcal{M}_P$, respectively, with the objectives detailed in the main text. During training, only $\mathcal{M}_P$ and the small MLP projector $h_\phi$ associated with $\mathcal{M}_F$ are trained while the others are kept frozen. During inference, both the foundational and prior models are discarded. 

\subsubsection{AR Generative Models.} We adopt Llama-like transformers~\cite{llama,llamagen} as our AR generative models. Following LARP~\cite{larp}, we leverage absolute learned positional encodings. During training, a dropout rate of 0.1 is applied to token sequences, residual connections, and feedforward layers. Furthermore, the SVQ quantizer of AVTok is configured to be deterministic during the training of AR generative models to encourage a more accurate latent representation learning. 

\subsection{Training Details}

\subsubsection{Reconstruction.} During the training of the AVTok tokenizer, $\mathcal{L}_{rec}^a$ and $\mathcal{L}_{rec}^v$ are the two primary objectives for the learning of audio and video streams. For video, $\mathcal{L}_{rec}^v$ comprises $L_1$ reconstruction term, LPIPS term~\cite{lpips} for perceptual enhancement, and GAN adversarial term~\cite{gan} for improved sharpness and fine-grained textual details, with corresponding weights of $(1.0, 1.0, 0.3)$ following~\cite{larp}. Similarly for audio, $\mathcal{L}_{rec}^a$ is a combination of Multi-Scale Mel-Spectrogram reconstruction term~\cite{dac}, deep feature matching term~\cite{hifigan}, and GAN adversarial term~\cite{bigvgan}, with respective weights of $(15.0, 2.0, 1.0)$ according to~\cite{bigvgan}. 

Notably, a ViT-based Discriminator~\cite{vit} is adopted to compute the GAN component of the video stream, while Multi-Scale Sub-Band CQT Discriminator~\cite{cqt} and Multi-Period Discriminator~\cite{hifigan} are employed to compute the GAN and feature matching components of the audio stream. These discriminators are updated once per five training iterations of the AVTok tokenizer with a 70\% lower learning rate and LeCam regularization~\cite{lecam} applied for training stability. Besides, SVQ quantization loss with total weight of 0.1 is also added, in which we follow~\cite{taming} to use a commitment and codebook loss weights of $(0.25, 1.0)$.

\begin{table}[tb]
  \caption{\textbf{Detailed training settings for the three stages of VFAL.}
  }
  \label{tab:vfal}
    \centering
    \renewcommand{\arraystretch}{1.25}
    \resizebox{1.\textwidth}{!}{
        \begin{tabular}{lccc}
        \toprule
        \textbf{Setting} & \textbf{Stage 1} & \textbf{Stage 2} & \textbf{Stage 3} \\ \midrule
        training purpose & Video Reconstruction & Audio Reconstruction & Refinement \\
        \multirow{3}{*}{trainable modules} & $\mathcal{E}(\cdot; LN_1^v, LN_2^v)$, & \multirow{2}{*}{$(LN_1^a, LN_2^a)_{\{\mathcal{E}, \mathcal{D}\}}$} & \multirow{3}{*}{$\mathcal{D}(\cdot;LN_1^{\{a,v\}},LN_2^{\{a,v\}})$} \\
        &$\mathcal{D}(\cdot; LN_1^v, LN_2^v)$ &\\
        &$\mathbf{Q}_L^v, \mathbf{Q}_P^v, \mathcal{Q}, \mathcal{M}_P$ &\multirow{-2}{*}{$\mathbf{Q}_L^a, \mathbf{Q}_P^a,\mathcal{M}_P, h_\phi$} \\
        base learning rate & 0.0001 & 0.0001 & 0.0001 \\
        scheduler & cosine & cosine & cosine \\
        $\beta_1, \beta_2$ & 0.9, 0.95 & 0.9, 0.95 & 0.9, 0.95 \\
        warm-up epochs & 8 & 3 & 1 \\
        total epochs & 75 & 35 & 10 \\
        batch size & 112 & 112 & 112 \\
        $\lambda_1,\lambda_2,\lambda_3,\lambda_4$ & 1.0, 0.0, 0.0, 0.06 & 0.1, 1.0, 0.5, 0.06 & 1.0, 0.01, 0.5, 0.06 \\
        \bottomrule
        \end{tabular}
        }
\end{table}

\begin{figure}[tb]
  \centering
  \includegraphics[width=1.\linewidth]{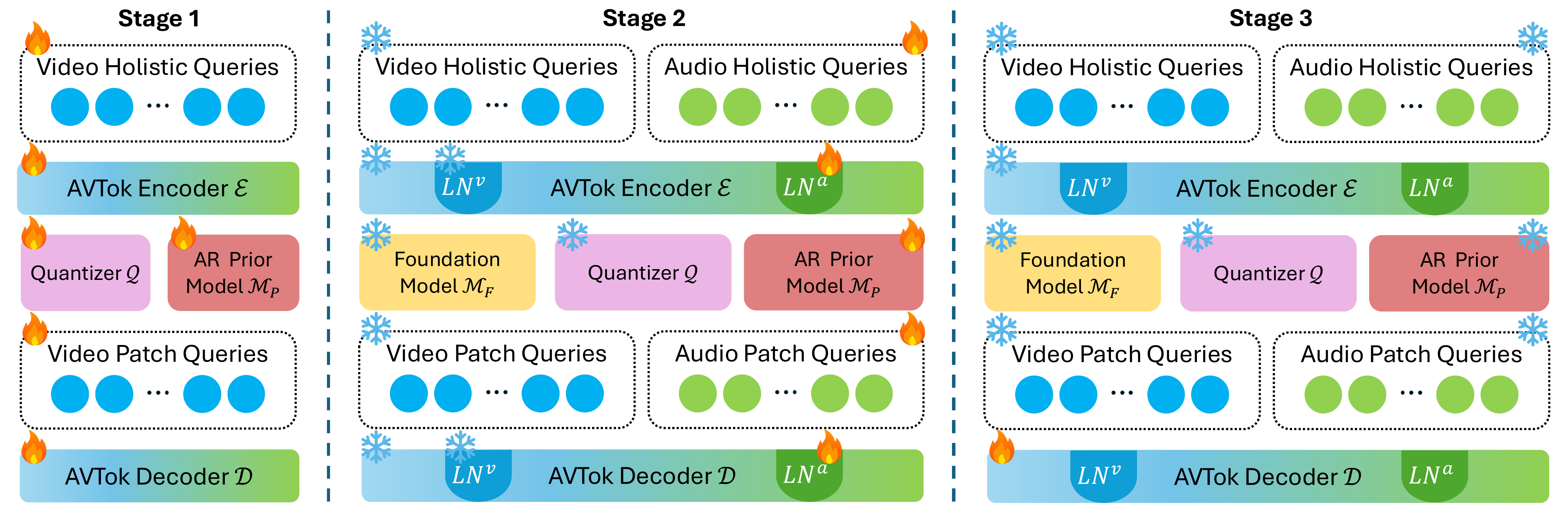}
  \caption{\textbf{Illustration of trainable parameters at different stages of VFAL.}
  }
  \label{fig:vfal}
\end{figure}

With the final training objective and the component weights $\lambda_{1,2,3,4}$ defined in the main text, we then conduct training for AVTok using the proposed VFAL hierarchical strategy. It is decomposed into three progressive stages with the primary target modules set as: (1) Video Reconstruction, when the encoder and decoder with video-specific normalization layers, denoted as $\mathcal{E}(\cdot;LN_1^v, LN_2^v)$ and $\mathcal{D}(\cdot;LN_1^v, LN_2^v)$, and learnable queries $(\mathbf{Q}_L^v, \mathbf{Q}_P^v)$ are trained for 75 epochs; (2) Audio Reconstruction, when $\mathcal{E},\mathcal{D}$ are frozen and shared between two streams except for audio-specific learnable queries $(\mathbf{Q}_L^a, \mathbf{Q}_P^a)$ and normalization layers $(LN_1^a, LN_2^a)_{\{\mathcal{E}, \mathcal{D}\}}$, which are trained for 35 epochs; (3) Refinement, when only the decoder with both streams $\mathcal{D}(\cdot;LN_1^{\{a,v\}},LN_2^{\{a,v\}})$ is further finetuned for 10 epochs. A batch size of $112$ and the Adam optimizer~\cite{adam} with a base $lr=0.0001$, $(\beta_1,\beta_2) = (0.9,0.95)$, and a warm-up cosine schedule are used for all stages. Additional details on other modules and training settings can be found in Tab.~\ref{tab:vfal} and Fig.~\ref{fig:vfal}. Note that in Stage 1, $\mathcal{L}_{prior}$ is computed with holistic video tokens $\mathbf{x}^v$ only. 

\subsubsection{Downstream Generation.} We train an AR generative model for each downstream generation task, including audio-to-video (A2V), video-to-audio (V2A), and class-conditional joint audio-video generation (cJAVG). They are all trained on the train split of VGGSound for 75 epochs with a batch size of 128. The AdamW optimizer~\cite{adamw} is used with $(\beta_1,\beta_2) = (0.9,0.95)$, a weight decay of $0.05$, and a base learning rate of $0.0006$, following a warm-up cosine learning rate schedule with $4$ warm-up epochs. When generating samples, we follow LARP~\cite{larp} to apply a small Classifier-Free Guidance (CFG) scale of 1.25~\cite{cfg} for cJAVG while excluding it for V2A and A2V tasks, and do not use top-k or top-p sampling methods. Besides, the conditioning video/audio for V2A/A2V is input to the corresponding streaming of AVTok to effortlessly produce the conditioning video/audio holistic tokens, while the \textit{vanilla} model may struggle. At inference time, the AR models predict the holistic tokens for the respective modality outcome, conditioned on the holistic tokens obtained.

\subsection{Evaluation Details}
\subsubsection{Representative Baselines.} For reconstruction, there is no open-source baseline available for direct comparison with our AVTok tokenizer due to the novelty of the proposed unified audio-video tokenization task. Therefore, we select several state-of-the-art unimodal 1D tokenizers of each modality that are closely relevant to AVTok for reasonable comparisons, as included in the main text. Regarding downstream generation, the baselines for each task are selected under a similar relevance consideration to ensure that the comparisons are as fair as possible. Since some baselines, such as Ovi~\cite{ovi}, JavisDiT~\cite{javisdit}, or MMAudio~\cite{mmaudio}, require textual captions as conditions to control the generation process, we bypass them by utilizing class labels available in VGGSound as an alternative. 
\subsubsection{Metrics.} To evaluate accuracy, realism, and perceptual quality of the reconstructed videos/audios with respect to the ground-truths, we respectively adopt PSNR~\cite{psnr}/SI-SDR~\cite{sisdr}, FVD~\cite{fvd}/FAD~\cite{fad}, and LPIPS~\cite{lpips}/MR-STFT~\cite{mrstft}. For downstream generation, we again use FVD~\cite{fvd}/FAD~\cite{fad} with additions of DeSync~\cite{syncformer} and ImageBind~\cite{imagebind} (IB) Score to assess realism, temporal synchronization, and semantic alignment accordingly. Notably, for A2V and V2A tasks, FVD/FAD are computed between the generated results and the ground-truths, while the rest are computed between the generated results and input conditions. For cJAVG, FAD and FVD are calculated similarly, whereas DeSync and IB Score are computed between the generated audio-video pairs.

\section{Additional Results}
\subsection{Ablation Study}

\begin{table}[tb]
  \caption{\textbf{Comparison of generation efficiency.}
  }
  \label{tab:latency}
  \centering
  \renewcommand{\arraystretch}{1.25}
    \resizebox{.75\textwidth}{!}{
        \begin{tabular}{c|r|c|c c|c|c}
            \Xhline{2.0\arrayrulewidth}
            \multirow{2}{*}{Task}& \multirow{2}{*}{Method}& Gen& \multicolumn{2}{c|}{\#Param}& Latency$\downarrow$ & \multirow{2}{*}{TFLOPs$\downarrow$}\\
            & & Type & Tokenizer & Generator & (sec) &\\
            \Xhline{2.0\arrayrulewidth}
             & TempoTokens~\cite{tempotoken} & Diff & 83.7M & 1.9B & 21.103 & 1.62K\\
             \multirow{-2}{*}{A2V} & \cellcolor{gray!10}{AVTok-A2V (Ours)} & \cellcolor{gray!10}{AR} & \cellcolor{gray!10}{208.4M} & \cellcolor{gray!10}{632.0M} & \cellcolor{gray!10}{11.058} & \cellcolor{gray!10}{1.82}\\
            \Xhline{2.0\arrayrulewidth}
            \multirow{5}{*}{V2A} & MMAudio~\cite{mmaudio} & FM & 298.5M & 1.3B & 1.304 & 31.77\\
            & VinTAGe~\cite{vintage} & FM & 110.6M & 1.5B & 23.423 & 474.69\\
            & V-AURA~\cite{v-aura} & AR & 76.7M & 816.9M & 11.290 & 191.05\\
            & SpecVQGAN~\cite{specvqgan} & AR & 76.4M & 332.4M & 1.307 & 17.72 \\
            & \cellcolor{gray!10}{AVTok-V2A (Ours)} & \cellcolor{gray!10}{AR} & \cellcolor{gray!10}{208.4M} & \cellcolor{gray!10}{632.0M} & \cellcolor{gray!10}{1.395} & \cellcolor{gray!10}{1.82}\\
            \Xhline{2.0\arrayrulewidth}
            \multirow{3}{*}{cJAVG} & JavisDiT~\cite{javisdit} & FM & 448.7M & 8.9B & 32.240 & 2.60K\\
            & Ovi~\cite{ovi} & FM & 988.6M & 17.3B & 87.282 & 14.99K\\ 
            & \cellcolor{gray!10}{AVTok-cJAVG (Ours)} & \cellcolor{gray!10}{AR} & \cellcolor{gray!10}{208.4M} & \cellcolor{gray!10}{632.4M} & \cellcolor{gray!10}{12.755} & \cellcolor{gray!10}{3.48}\\
            \Xhline{2.0\arrayrulewidth}
        \end{tabular}
    }
\end{table}
\begin{table}[tb]
  \caption{\textbf{Comparison of different model scale.}
  }
  \label{tab:scalability}
  \centering
  \renewcommand{\arraystretch}{1.25}
    \resizebox{1.\textwidth}{!}{
        \begin{tabular}{l|c c c|p{14mm}<{\centering} p{14mm}<{\centering} p{14mm}<{\centering}|p{14mm}<{\centering} p{14mm}<{\centering} p{17mm}<{\centering}}
            \Xhline{2.0\arrayrulewidth}
            \multirow{2}{*}{Model}& \multicolumn{3}{c|}{Configuration}& \multicolumn{3}{c|}{Video Reconstruction}& \multicolumn{3}{c}{Audio Reconstruction} \\
            & Hidden Size & Depth & Num Heads & PSNR$\uparrow$ & rFVD$\downarrow$ & LPIPS$\downarrow$ & SI-SDR$\downarrow$ & rFAD$\downarrow$ & MR-STFT$\downarrow$ \\
            \Xhline{2.0\arrayrulewidth}
            \rowcolor{gray!10} AVTok & 768 & 12 & 12 & 25.62 & 12.80 & 0.126 & 23.09 & 5.93 & 1.523\\
            AVTok-B & 768 & 8 & 12 & 24.65 & 12.94 & 0.148 & 24.99 & 6.01 & 1.794\\
            AVTok-S & 768 & 6 & 8 & 23.39 & 19.12 & 0.193 & 25.29 & 8.86 & 2.333\\
            \Xhline{2.0\arrayrulewidth}
        \end{tabular}
    }
\end{table}
\subsubsection{Generation Latency.} In addition to model capacity measured by the number of parameters, we evaluate the efficiency of our complete generation pipeline comprising AVTok tokenizer and an AR generative model compared to the other baselines. Specifically, we measure the TFLOPs and average latency per sample of all methods in generating $100$ samples with a batch size of $1$ in the same environment. For other settings of the baselines, we use their default configuration as deemed necessary. The results shown in Tab.~\ref{tab:latency} highlight the efficiency of our pipelines, complementing their effectiveness to generate high-fidelity samples as demonstrated in the main text.

\subsubsection{Tokenizer Scalability.}  
To explore the effect of scaling our AVTok tokenizer, we adjust the model size while maintaining the same number of latent tokens to construct another two smaller variants AVTok-B and AVTok-S, and conduct training for them under identical settings as the default. As shown in Tab.~\ref{tab:scalability}, the performance changes most significantly when scaling from AVTok-S up to AVTok-B with 6.18 FVD and 2.85 FAD lower results, but saturates with minor improvements when scaling up to the largest default. Given this indication, we opt not to scale the model further and use the current largest variant as the default choice.

In addition, we also examine the model performance when adjusting the number of holistic tokens necessary to encode and reconstruct input audio and video. In particular, we alternately halve the default number of tokens for one modality while keeping that of the other unchanged to quantify the effect they induce.
Intuitively, the use of fewer tokens enables a faster AR generation process but trades off with degradation in reconstruction quality, which is reflected in Tab.~\ref{tab:tok_count}. Interestingly, we find that a decreasing number of video tokens significantly affects audio reconstruction, while a marginal impact is observed conversely. This suggests that the model is more susceptible to changes in the video stream, where the information is denser and richer than the audio stream.

\subsubsection{External Models.}
We also ablate on different selections of external models, including the vocoders and audio-visual foundation models $\mathcal{M}_F$. First, we replace CAV-MAE Sync~\cite{cavmae-plus} with its predecessor CAV-MAE~\cite{cavmae} as our foundational model, which results in performance degradation despite having similar model size. Second, we adopt BigVGAN~\cite{bigvgan}, a more robust neural vocoder, as an alternative to HiFi-GAN~\cite{hifigan}, which only yields slight improvements and has a significantly larger model size. Therefore, we opt to use CAV-MAE Sync~\cite{cavmae-plus} and BigVGAN~\cite{bigvgan} by default, considering the balance of efficiency and effectiveness.

\begin{table}[tb]
  \caption{\textbf{Comparison of different holistic token counts.}
  }
  \label{tab:tok_count}
  \centering
  \renewcommand{\arraystretch}{1.25}
    \resizebox{1.\textwidth}{!}{
        \begin{tabular}{l|c c|p{14mm}<{\centering} p{14mm}<{\centering} p{14mm}<{\centering}|p{14mm}<{\centering} p{14mm}<{\centering} p{17mm}<{\centering}}
            \Xhline{2.0\arrayrulewidth}
            \multirow{2}{*}{Model}& \multicolumn{2}{c|}{Configuration}& \multicolumn{3}{c|}{Video Reconstruction}& \multicolumn{3}{c}{Audio Reconstruction} \\
            & \#Video Tokens & \#Audio Tokens & PSNR$\uparrow$ & rFVD$\downarrow$ & LPIPS$\downarrow$ & SI-SDR$\downarrow$ & rFAD$\downarrow$ & MR-STFT$\downarrow$ \\
            \Xhline{2.0\arrayrulewidth}
            \rowcolor{gray!10} AVTok & 1024 & 128 & 25.62 & 12.80 & 0.126 & 23.09& 5.93 & 1.523\\
            AVTok-a64 & 1024 & 64 & 25.37 & 12.75 & 0.128 & 25.35 & 12.77 & 2.263 \\
            AVTok-v512 & 512 & 128 & 23.94 & 23.85 & 0.172 & 26.10 & 14.90 & 2.491\\
            \Xhline{2.0\arrayrulewidth}
        \end{tabular}
    }
\end{table}

\begin{table}[tb]
  \caption{\textbf{Comparison of using different foundational models and vocoders.}
  }
  \label{tab:fm_vo}
  \centering
  \renewcommand{\arraystretch}{1.25}
    \resizebox{1.\textwidth}{!}{
        \begin{tabular}{l|r r|p{14mm}<{\centering} p{14mm}<{\centering} p{14mm}<{\centering}|p{14mm}<{\centering} p{14mm}<{\centering} p{17mm}<{\centering}}
            \Xhline{2.0\arrayrulewidth}
            \multirow{2}{*}{Model}& \multicolumn{2}{c|}{Configuration}& \multicolumn{3}{c|}{Video Reconstruction}& \multicolumn{3}{c}{Audio Reconstruction} \\
            & \multicolumn{1}{c}{$\mathcal{M}_F$} & \multicolumn{1}{c|}{Vocoder} & PSNR$\uparrow$ & rFVD$\downarrow$ & LPIPS$\downarrow$ & SI-SDR$\downarrow$ & rFAD$\downarrow$ & MR-STFT$\downarrow$ \\
            \Xhline{2.0\arrayrulewidth}
            \rowcolor{gray!10} AVTok & CAV-MAE Sync~\cite{cavmae-plus} & HiFi-GAN~\cite{hifigan} & 25.62 & 12.80 & 0.126 & 23.09& 5.93 & 1.523\\
            AVTok-F & CAV-MAE~\cite{cavmae} & HiFi-GAN~\cite{hifigan} & 25.50 & 12.84 & 0.128 & 24.19 & 6.40 & 1.622\\
            AVTok-V & CAV-MAE Sync~\cite{cavmae-plus} & BigVGAN~\cite{bigvgan} & 25.61 & 12.59 & 0.125 & 22.78 & 5.72 & 1.511\\
            \Xhline{2.0\arrayrulewidth}
        \end{tabular}
    }
\end{table}

\subsection{Visualization}
\begin{figure}[tb]
  \centering
  \includegraphics[width=1.\linewidth]{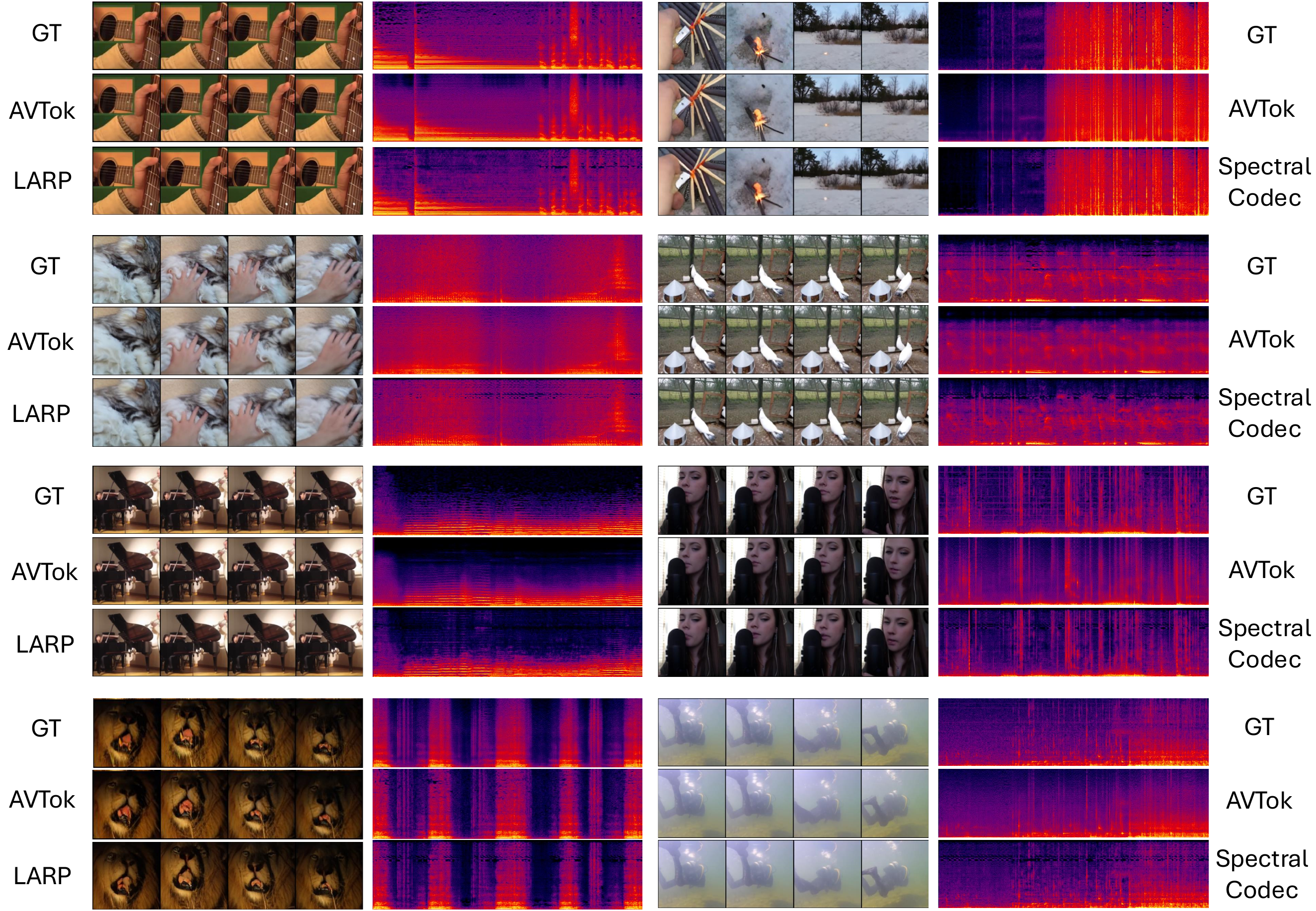}
  \caption{\textbf{Additional qualitative reconstruction results.}
  }
  \label{fig:sup_rec}
\end{figure}

\begin{figure}[tb]
  \centering
  \includegraphics[width=1.\linewidth]{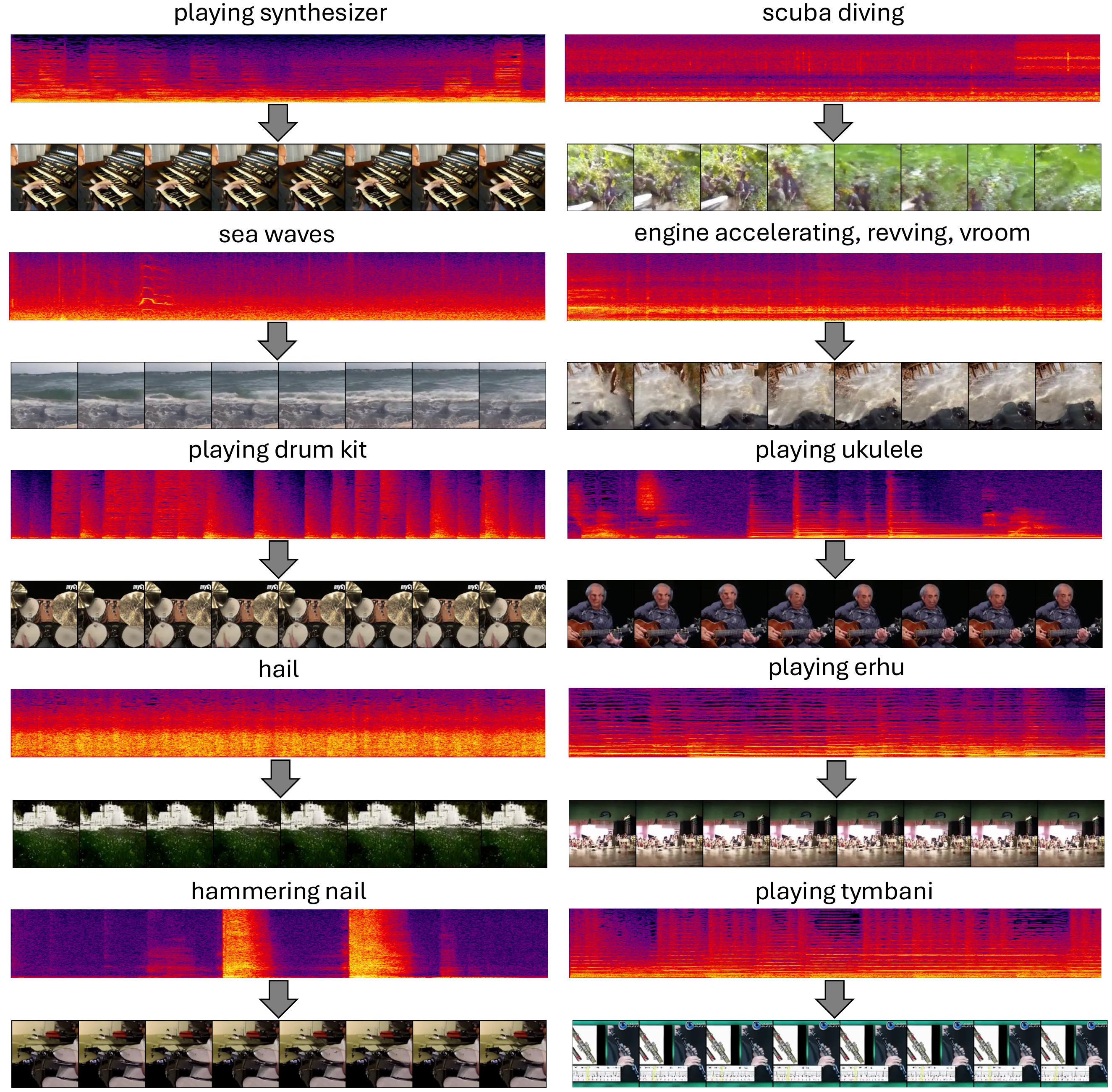}
  \caption{\textbf{Additional qualitative results for audio-to-video generation task.}
  }
  \label{fig:sup_a2v}
\end{figure}

\begin{figure}[tb]
  \centering
  \includegraphics[width=1.\linewidth]{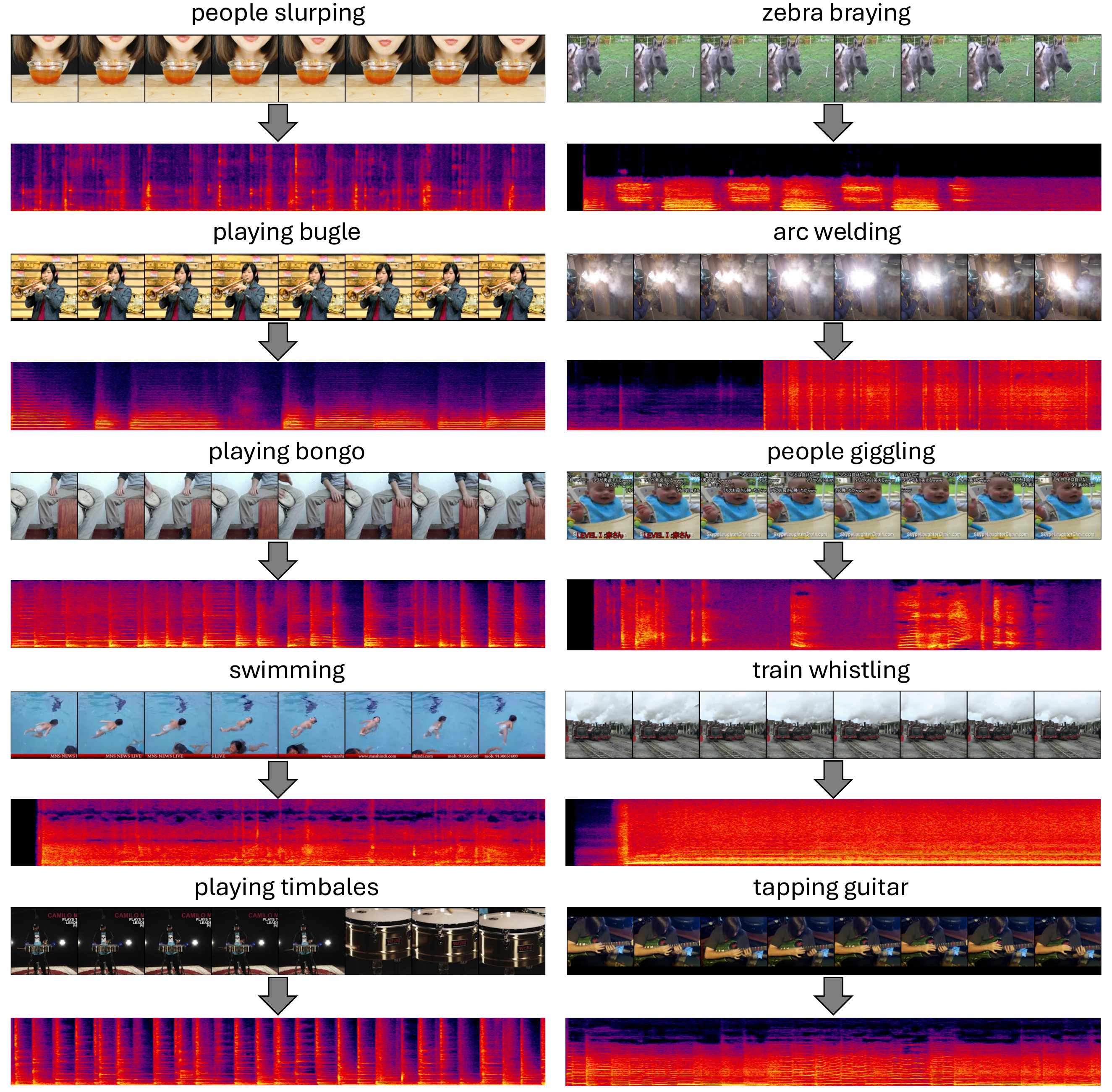}
  \caption{\textbf{Additional qualitative results for video-to-audio generation task.}
  }
  \label{fig:sup_v2a}
\end{figure}

\begin{figure}[tb]
  \centering
  \includegraphics[width=1.\linewidth]{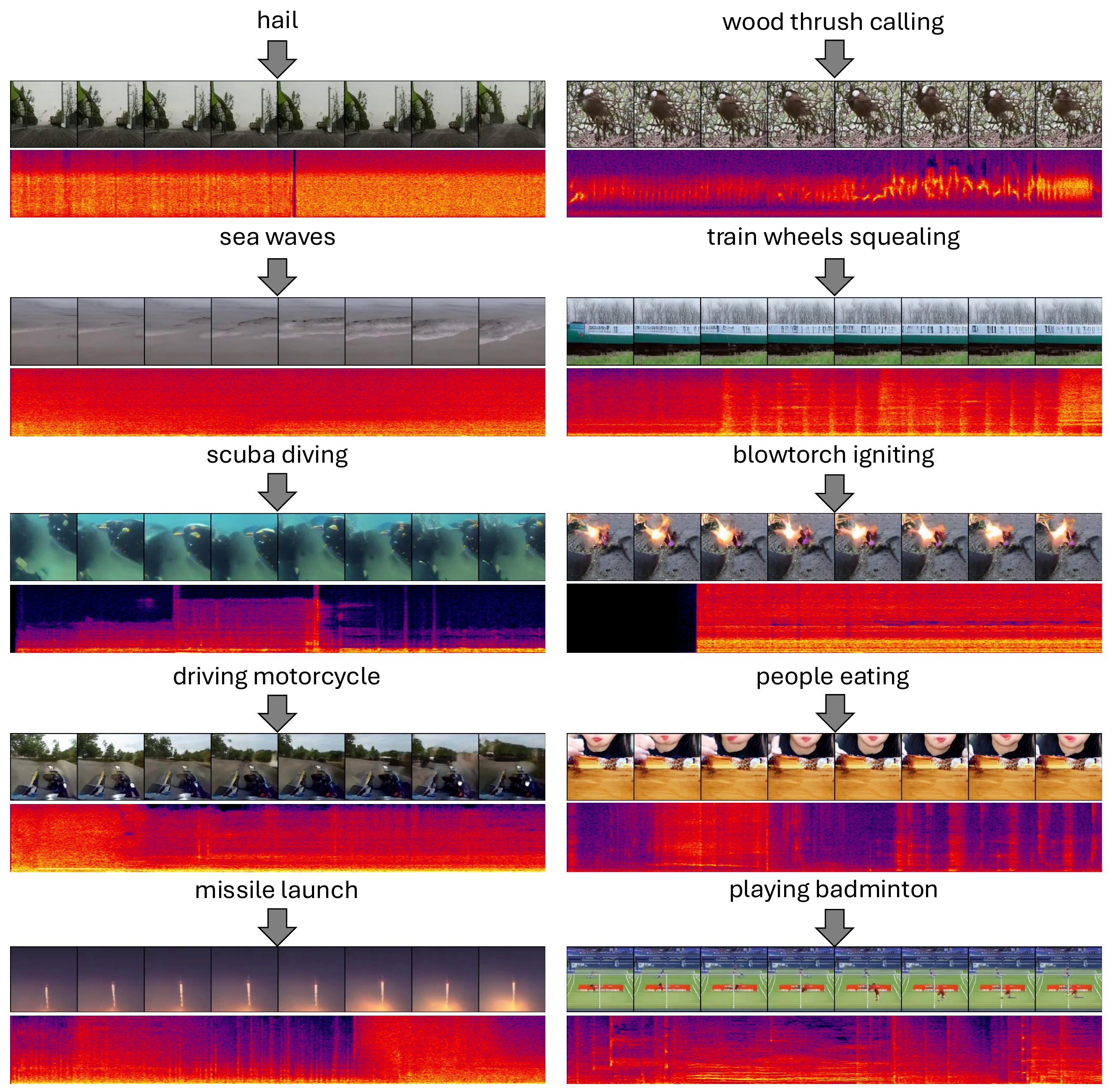}
  \caption{\textbf{Additional qualitative results for class-conditional joint generation task.}
  }
  \label{fig:sup_cjavg}
\end{figure}

We provide additional qualitative results for reconstruction, audio-to-video, video-to-audio, and class-conditional joint audio-video generation tasks in Fig.~\ref{fig:sup_rec},~\ref{fig:sup_a2v},~\ref{fig:sup_v2a},~\ref{fig:sup_cjavg}, respectively. These results consistently highlight that the AVTok tokenizer excels in both reconstruction and when incorporated into the downstream AR generative models for generation tasks. Besides, generated sounding video files in MP4 format are also included for subjective inspection.

\section{Discussion}
\subsection{Potential Limitations}
Our proposed AVTok unified tokenizer demonstrates outstanding performance in audio-video tokenization, and excels when integrated into our AR generative models for downstream generation tasks. However, certain limitations remain, which open promising directions for future exploration.

\subsubsection{Data Scale and Complexity.} Our AVTok tokenizer and AR generative models are trained on approximately 640K and 180K data entries only. Despite maintaining efficacy and efficiency, this may inevitably constrain scalability compared to larger proprietary models and systems. In addition, due to the inherent simplicity of the scenes included in the datasets, artifacts may appear in the generated samples when scenes are particularly complex. We believe that larger-scale training with more diverse and high-quality audio-video datasets could enhance the robustness and generalizability of the models.

\subsubsection{Model Design and Training Resource.} Similar to other transformer-based unimodal tokenizers, AVTok inherently performs best with fixed-resolution audios and videos due to positional encoding constraints. Besides, amid limited training resources, we could only train and evaluate our models' capabilities on 16-frame short clips with $128\times128\times3$ low resolution and roughly 4-second audios with a sampling rate of 22kHz. We contemplate that with sufficient resources, scaling up AVTok and the AR generative models can enable reconstructing and synthesizing audio-video pairs with larger resolution, longer duration, and higher quality, meeting user demands nowadays.

\subsubsection{Synchronization Modeling.} The current architecture and training setups of AVTok tokenizer and AR generative models only partially exploit synchronization between auditory and visual elements in an implicit manner by: (1) using synchronized sounding video samples as input; and (2) enabling cross-modal interaction with shared parameters, AR prior and foundation models for AVTok, and causal self-attention mechanism in AR generative models. Therefore, the generated audios and videos may lack temporal alignment. We think that modeling synchronization between the two modalities more explicitly can help mitigate this issue and improve the final performance. 

\subsubsection{End-to-end Training.} The training of the AVTok tokenizer relies on the proposed VFAL hierarchical strategy. Although effective, it requires complicated stage-wise tuning and is prone to cascading errors, which may eventually lead to suboptimal performance. Conversely, a single-stage end-to-end alternative could alleviate these issues by streamlining the training process with a unified objective, albeit with higher optimization sensitivity and computational cost.

\subsection{Statements}
\subsubsection{Ethics.} All datasets and models used in this work are publicly accessible online and contain no private or sensitive information.
\subsubsection{Reproducibility.} To ensure full reproducibility, we detail our model’s design, training, and evaluation in the main text and appendix, and will publicly release all code, checkpoints, and datasets.

\subsubsection{LLM Usage.} Large Language Models (LLMs) were utilized solely as writing aids to polish the language and refine the presentation. They played no role in developing the core concepts or research design. 


%
%
\end{document}